\documentclass[10pt,twocolumn,letterpaper]{article}
\usepackage[pagenumbers]{cvpr}
\usepackage{bm}
\usepackage{mathtools}
\usepackage{amsthm}
\usepackage{microtype}
\usepackage{booktabs,graphicx,makecell,colortbl}
\usepackage{tikz}
\usepackage{float}
\usetikzlibrary{arrows.meta,calc}
\definecolor{cvprblue}{rgb}{0.21,0.49,0.74}
\usepackage[pagebackref,breaklinks,colorlinks,allcolors=cvprblue]{hyperref}
\makeatletter
\g@addto@macro\normalsize{%
  \setlength\abovedisplayskip{4pt plus 2pt minus 2pt}%
  \setlength\belowdisplayskip{4pt plus 2pt minus 2pt}%
  \setlength\abovedisplayshortskip{0pt plus 2pt}%
  \setlength\belowdisplayshortskip{2pt plus 2pt minus 1pt}}
\makeatother
\usetikzlibrary{decorations.pathreplacing}
\definecolor{accent}{rgb}{0.85,0.45,0.10}
\definecolor{ink}{rgb}{0.20,0.22,0.27}
\definecolor{revisionpurple}{RGB}{128,0,128}

\newcommand{\R}{\mathbb{R}}
\newcommand{\E}{\mathbb{E}}
\newcommand{\method}{TRQ}
\providecommand{\lbltbl}[1]{\label{tab:#1}}
\DeclareMathOperator*{\argmax}{arg\,max}
\theoremstyle{plain}

\newtheorem{rotationtheorem}{Theorem}
\newtheorem{rotationcorollary}{Corollary}
\title{Online Temporal Rotations for 2 Bit KV-Cache Quantization\\in Video Diffusion Transformers}
\author{Ryan Lucas\hspace{2em}%
Petros Prastakos\hspace{2em}%
Rahul Mazumder\\[5pt]
{\normalsize Operations Research Center,
Massachusetts Institute of Technology}\\[2pt]
{\small\ttfamily \{ryanlu,pprastak,rahulmaz\}@mit.edu}}
\hypersetup{
  pdftitle={Online Temporal Rotations for 2 Bit KV-Cache Quantization in Video Diffusion Transformers},
  pdfauthor={Ryan Lucas, Petros Prastakos, Rahul Mazumder}
}
\makeatletter
\newcommand{\compactparagraph}{\@startsection{paragraph}{4}{\z@}
  {6pt plus 1pt minus 1pt}{-1em}{\normalfont\normalsize\bfseries}}
\let\trqOriginalSection\cvprsection
\let\trqOriginalSubsection\cvprsubsection
\newcommand{\trqUseCompactHeadingSpacing}{%
\patchcmd{\cvprsection}{-10pt plus -2pt minus -2pt}{-8pt plus -1pt minus -1pt}
  {}{\PackageError{trq-spacing}{Section spacing patch failed}{Check the cvpr.sty version.}}
\patchcmd{\cvprsection}{7pt}{5pt}
  {}{\PackageError{trq-spacing}{Section spacing patch failed}{Check the cvpr.sty version.}}
\patchcmd{\cvprsubsection}{-8pt plus -2pt minus -2pt}{-6pt plus -1pt minus -1pt}
  {}{\PackageError{trq-spacing}{Subsection spacing patch failed}{Check the cvpr.sty version.}}
\patchcmd{\cvprsubsection}{5pt}{4pt}
  {}{\PackageError{trq-spacing}{Subsection spacing patch failed}{Check the cvpr.sty version.}}
}
\newcommand{\trqRestoreHeadingSpacing}{%
  \let\cvprsection\trqOriginalSection
  \let\cvprsubsection\trqOriginalSubsection}
\makeatother
\begin{document}
\maketitle

\begin{abstract}
Video diffusion transformers reuse key--value (KV) caches across denoising steps, which quickly becomes a bottleneck for long video generation. We find that quantizing temporally distant cache vectors leads to a large range and high quantization error, but that nearby video tokens in the KV-cache share a persistent pattern per channel. These patterns also vary across videos, motivating rotations that are adapted to each video's unique cache. We propose Temporal Rotation Quantization (TRQ), a method that rotates frames in the cache that are adjacent in time before quantization. For each block of temporally adjacent cache vectors, TRQ fits an orthogonal transform that maps the block's per-channel means to a vector that is close to a constant across channels. To keep the quantization format lightweight, we parameterize this rotation as a tree of $2 \times 2$ planar rotations, each stored with only two bits, followed by a fixed Hadamard transform shared across all blocks. The rotations are fit without gradients or changes to model weights, and we prove that a simple greedy fit is globally optimal for our 'mean-flattening' objective. The transforms are fit online for each new video, and we release fused Triton kernels to make this process efficient at inference. At 2-bit precision and matched compression, TRQ improves VBench subject consistency over the state-of-the-art method QVG from 90.38 to 92.02 on LongCat-Video-13B and from 86.33 to 87.61 on HY-WorldPlay-8B, as well as improving other metrics. 
\end{abstract}
\vspace{-5mm}
\section{Introduction}
\label{sec:intro}
Diffusion transformers (DiTs)~\cite{dit} have become the dominant architecture for video generation. They underpin proprietary systems such as Sora~\cite{sora} and Movie Gen~\cite{moviegen}, as well as open models including CogVideoX~\cite{cogvideox}, HunyuanVideo~\cite{hunyuanvideo}, and Wan~\cite{wan}, which synthesize high-fidelity videos from text or image prompts. To go beyond a few seconds of footage, toward minute-scale or interactive generation, recent work produces video autoregressively or by continuation, denoising one chunk at a time conditioned on previously generated content~\cite{diffusionforcing,causvid,selfforcing,magi1,skyreels2,longcat}. Like autoregressive language models, these systems store the keys and values of past frames in a KV-cache, so the history need not be recomputed at every denoising step.

However, this computational saving from the KV-cache brings with it substantial memory requirements. Each frame contributes thousands of tokens, and the cache holds a key and a value vector per token, head, and layer. It therefore grows linearly with the generated history and can reach tens of gigabytes, more than commodity GPUs can hold~\cite{qvg}. Capping the cache, for example with a sliding window, shrinks the context the model can attend to and erodes long-range consistency of subjects, scene layout, and motion~\cite{qvg}. KV-cache quantization is mature for large language models~\cite{kivi,kvquant,quarot}, but these methods were designed for the unique structure of LLM KV-caches and lose substantial accuracy when applied directly to video DiTs~\cite{qvg}. 

% We study this problem in LongCat-Video~\cite{longcat}, which encodes a clean conditioning prefix once into a KV-cache and reuses it across all denoising steps of each continuation chunk.

\begin{figure*}[t]
\centering
\includegraphics[width=0.95\textwidth]{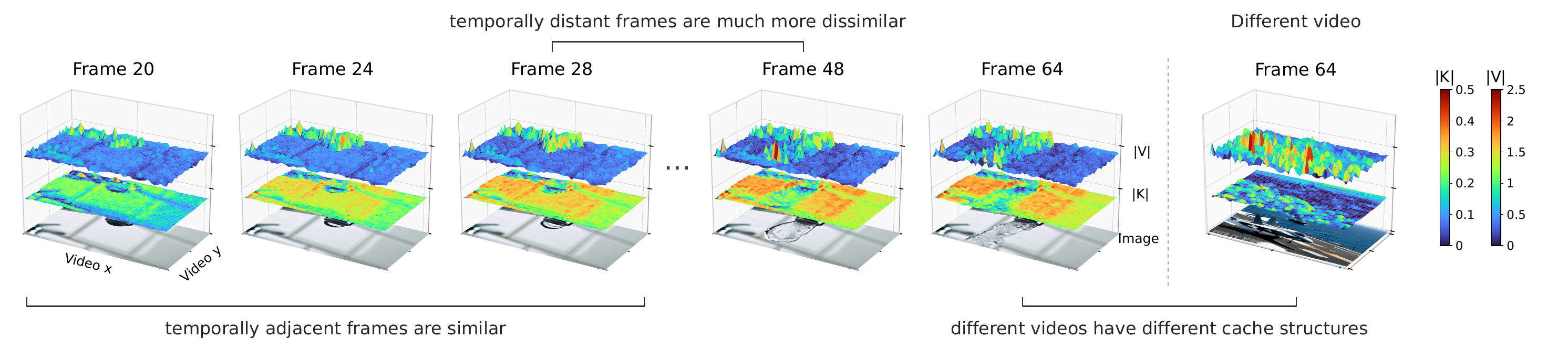}
\caption{Structure of video-diffusion KV cache. Each panel overlays key and value magnitudes ($|\bm{K}|$ and $|\bm{V}|$) on the corresponding video frame. We observe that (i) Within a video, nearby frames  share persistent spatial patterns and (ii) the final two panels show that these patterns differ across videos, even at the same frame index and channel. Together, these observations motivate fitting rotations online to each video's cache and sharing them within local temporal blocks. For other channels, see Section~7 of the appendix.}
\label{fig:kv-cache}
\end{figure*}
% Figure 1 uses layer 0, head 0, feature-Haar coefficient 0 and absolute values.
% Source frames 20, 24, 28, 48, 64 correspond to latent indices 0, 1, 2, 7, 11.

Rotation transforms have proved remarkably effective for low-bit LLM quantization~\cite{quarot,spinquant,rotatekv, thinquant}. A central motivation for rotations in KV-cache quantization of LLMs is the empirically well-documented outlier structure along the channel dimension of the activations and KV-cache of LLMs. A small set of channels contain consistently large values across all tokens, forcing low-bit quantization to cover a wide range at the expense of resolution for smaller values. One intuition for why rotations are so effective in LLMs is that they take linear combinations of channels, `mixing' the outlier channels with more benign channels and thus leading to much flatter distributions that are more friendly to quantization~\cite{quarot}.

We identify two unique properties of video diffusion KV-caches that motivate online temporal rotations (Figure~\ref{fig:kv-cache}).

\textbf{(i) Temporal structure within a video.} For a given video and channel of the cache, keys and values are highly correlated across nearby frames, with only localized changes as the scene evolves. We illustrate this structure in the first five panels of Figure~\ref{fig:kv-cache}. This similarity, however, decays over time. Frames that are distant in time, such as frames 20 and 64 of the first video, can have substantially different values.

\textbf{(ii) Heterogeneous structure across videos.} Although the cache is stable over adjacent frames within a video, the pattern itself differs from one video to the next. The final two panels of Figure~\ref{fig:kv-cache} show frame 64 of two different videos at the same layer, head, and channel, and their key and value patterns are markedly different.

\begin{figure}[H]
\centering
\resizebox{.85\columnwidth}{!}{%
\begin{tikzpicture}[
    font=\sffamily,
    >={Stealth[length=3.2pt,width=2.6pt]},
    frame/.style={draw=cvprblue!55, line width=0.4pt, fill=cvprblue!7, rounded corners=0.6pt},
    patch/.style={draw=accent, line width=0.5pt, fill=accent!35},
    cell/.style={draw=white, line width=0.35pt, minimum width=0.15cm,
                 minimum height=0.13cm, inner sep=0pt},
    blockbg/.style={fill=#1, rounded corners=4pt, draw=none},
    rot/.style={draw=accent!80!black, line width=0.5pt, fill=white,
                rounded corners=3pt, inner xsep=3pt, inner ysep=2.5pt,
                font=\footnotesize\color{ink}},
    rotK/.style={rot, draw=cvprblue!85!black, fill=cvprblue!4, font=\fontsize{6.5}{7}\selectfont\color{ink}, inner xsep=2.5pt, inner ysep=1.5pt},
    rotV/.style={rot, draw=accent!85!black, fill=accent!4, font=\fontsize{6.5}{7}\selectfont\color{ink}, inner xsep=2.5pt, inner ysep=1.5pt},
    wireK/.style={cvprblue!80!black, line width=0.45pt},
    wireV/.style={accent!80!black, line width=0.45pt},
    lbl/.style={font=\scriptsize\sffamily, text=black!55},
]
\def\dx{0.66}\def\gap{0.22}
% ---------- block backgrounds ----------
\foreach \b/\col in {0/cvprblue!6, 1/cvprblue!6, 2/cvprblue!6} {
  \pgfmathsetmacro{\o}{\b*(3*\dx+\gap)}
  \fill[blockbg=\col] (\o-0.30,-2.52) rectangle (\o+2*\dx+0.52,0.74);
}
% ---------- video frames (slanted tiles with an evolving scene) ----------
\foreach \t in {0,...,8} {
  \pgfmathsetmacro{\x}{\t*\dx+int(\t/3)*\gap}
  \begin{scope}[shift={(\x,0)}]
    \fill[cvprblue!5] (-0.22,0.00) -- (0.30,0.00) -- (0.44,0.48) -- (-0.08,0.48) -- cycle;
    \begin{scope}
      \clip (-0.22,0.00) -- (0.30,0.00) -- (0.44,0.48) -- (-0.08,0.48) -- cycle;
      % ground with a slowly tilting horizon (camera motion)
      \pgfmathsetmacro{\hl}{0.13+0.012*\t}
      \pgfmathsetmacro{\hr}{0.21-0.010*\t}
      \fill[cvprblue!14] (-0.3,-0.1) -- (-0.3,\hl) -- (0.5,\hr) -- (0.5,-0.1) -- cycle;
      % a ball arcing across the scene, passing over the tracked patch
      \pgfmathsetmacro{\bx}{-0.10+0.055*\t}
      \pgfmathsetmacro{\by}{0.14+0.026*abs(\t-4)}
      \fill[cvprblue!45] (\bx,\by) circle (0.062);
      % a tree entering from the right
      \pgfmathsetmacro{\tx}{0.62-0.07*\t}
      \fill[cvprblue!32] (\tx-0.055,0.16) -- (\tx+0.055,0.16) -- (\tx,0.36) -- cycle;
      \fill[cvprblue!40] (\tx-0.008,0.10) rectangle (\tx+0.008,0.16);
    \end{scope}
    \draw[frame, fill=none] (-0.22,0.00) -- (0.30,0.00) -- (0.44,0.48) -- (-0.08,0.48) -- cycle;
    % tracked image patch (same location in every frame)
    \draw[patch, fill opacity=0.45] (0.10,0.08) -- (0.20,0.08) -- (0.23,0.18) -- (0.13,0.18) -- cycle;
    \draw[accent!60, line width=0.35pt, densely dotted] (0.165,0.08) -- (0.055,-0.30);
  \end{scope}
}
% ---------- cached K / V vectors for the tracked patch ----------
% per-block channel means (5 channels), shared by frames of that block
\def\meansK{{{70,25,55,10,40},{30,65,15,50,20},{55,35,70,20,60}}}
\def\meansV{{{20,60,35,70,15},{60,20,45,15,65},{25,50,20,65,40}}}
\pgfmathsetseed{7}
\foreach \t in {0,...,8} {
  \pgfmathsetmacro{\x}{\t*\dx+int(\t/3)*\gap}
  \pgfmathtruncatemacro{\b}{int(\t/3)}
  \foreach \c in {0,...,4} {
    \pgfmathsetmacro{\y}{-0.40-\c*0.135}
    \pgfmathtruncatemacro{\vk}{min(95,max(4,\meansK[\b][\c]+rand*5))}
    \pgfmathtruncatemacro{\vv}{min(95,max(4,\meansV[\b][\c]+rand*5))}
    \node[cell, fill=cvprblue!\vk!white] at (\x-0.045,\y) {};
    \node[cell, fill=accent!\vv!white]   at (\x+0.155,\y) {};
  }
}
% K / V labels
\node[lbl, anchor=east] at (-0.34,-0.67) {\begin{tabular}{@{}r@{}}cache\\[-1pt]$K$\,{\color{cvprblue}\rule{3pt}{3pt}}\;$V$\,{\color{accent}\rule{3pt}{3pt}}\end{tabular}};

\node[lbl, anchor=east] at (-0.34,0.24) {frames};
% ---------- separate K and V rotations per block ----------
% K bus sits above the V bus; K enters from the left, V from the right, and the
% only crossings (V drops over the K bus) are drawn as explicit hops.
\def\yK{-1.27}\def\yV{-1.48}\def\hop{0.055}
\foreach \b in {0,1,2} {
  \pgfmathsetmacro{\o}{\b*(3*\dx+\gap)}\pgfmathsetmacro{\cx}{\o+\dx+0.055}
  \pgfmathtruncatemacro{\n}{\b+1}
  \node[rotK] (K\b) at (\cx-0.40,-1.84) {$R^{K}_{B_{\n}}$};
  \node[rotV] (V\b) at (\cx+0.40,-1.84) {$R^{V}_{B_{\n}}$};
  % key wiring: elbow from the badge's left side into the K bus
  \draw[wireK, rounded corners=2pt] (K\b.west) -| (\o-0.17,\yK) -- (\o+2*\dx-0.045,\yK);
  \foreach \k in {0,1,2} {
    \draw[wireK,->] (\o+\k*\dx-0.045,\yK) -- (\o+\k*\dx-0.045,-1.07);
    \fill[cvprblue!80!black] (\o+\k*\dx-0.045,\yK) circle (0.7pt);
  }
  % value wiring: elbow from the badge's right side into the V bus
  \draw[wireV, rounded corners=2pt] (V\b.east) -| (\o+2*\dx+0.30,\yV) -- (\o+0.155,\yV);
  \foreach \k in {0,1,2} {
    \pgfmathsetmacro{\vx}{\o+\k*\dx+0.155}
    \ifnum\k<2
      % hop over the K bus
      \draw[wireV,->,preaction={draw=cvprblue!6,line width=2pt}] (\vx,\yV) -- (\vx,\yK-\hop) arc[start angle=-90, end angle=90, radius=\hop] -- (\vx,-1.07);
    \else
      \draw[wireV,->] (\vx,\yV) -- (\vx,-1.07);
    \fi
    \fill[accent!80!black] (\vx,\yV) circle (0.7pt);
  }
  % brace + block label
  \draw[decorate, decoration={brace, mirror, amplitude=3pt}, black!35, line width=0.4pt]
    (\o-0.16,-2.12) -- (\o+2*\dx+0.27,-2.12);
  \node[lbl] at (\cx,-2.34) {block $B_{\n}$};
}
% ---------- time axis ----------
\draw[->, black!45, line width=0.5pt] (-0.30,-2.68) -- ({8*\dx+2*\gap+0.70},-2.68)
  node[right, lbl] {$t$};
\node[lbl, anchor=north] at ({4*\dx+\gap+0.14},-2.75)
  {time};
\end{tikzpicture}%
}
\caption{Block-wise temporal rotations. At each image patch, the cached keys and
values are divided into temporal blocks of $n_B$ consecutive latent frames. Each
block receives its own key rotation $R^{K}_{B}$ and value rotation $R^{V}_{B}$, applied to
every frame in the block.}
\label{fig:blockwise-rotation}
\end{figure}
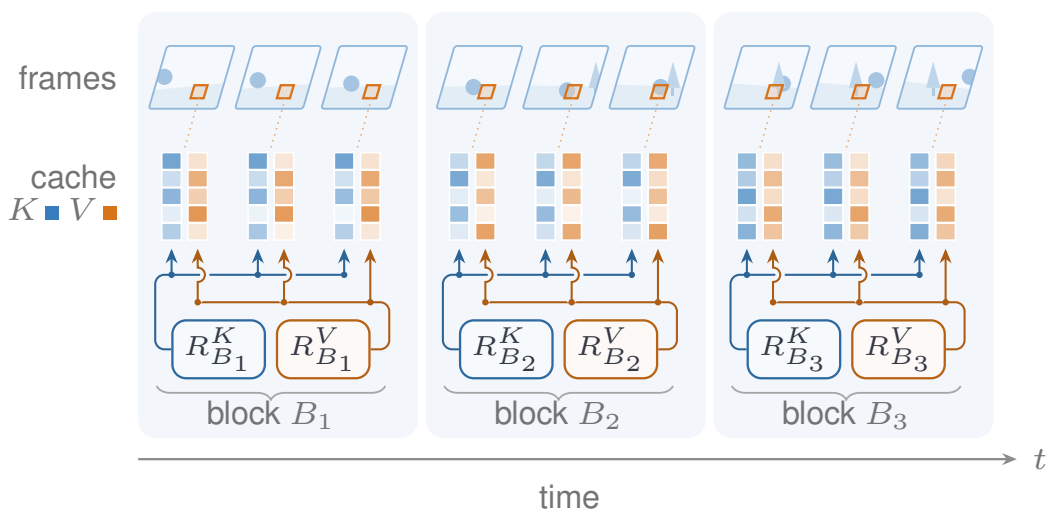

A generic channel rotation, such as the fixed random rotations used by QuaRot~\cite{quarot}, does not explicitly exploit either of these properties. A fixed rotation must simultaneously accommodate temporal and cross-channel variation with a single rotation. Moreover, because this temporal structure differs across videos, a rotation chosen before generation cannot adapt to the cache of the video actually being generated. Instead of applying a single rotation shared across all tokens, TRQ allows adjacent frames \emph{within one video} to share a rotation adapted to the content they have in common. Specifically, we divide the cache at each patch into  blocks of $n_B$ consecutive latent frames and fit one orthogonal rotation $\bm{R}_B$ per block, which is applied to every frame in the block (Fig.~\ref{fig:blockwise-rotation}). Fitting the rotation separately for each video and block allows it to track the scene as it evolves, while sharing the rotation within a block amortizes its storage cost over $n_B$ cache vectors.

% We find that the KV-cache of video DiTs is dominated by a \emph{local temporal mean} rather than only by isolated outlier channels. At a given image patch, the cached vectors of nearby frames share a common component that varies across channels but changes slowly over time, together with small frame-specific deviations. This structure suggests a different use of rotations than in LLMs. Instead of applying a single rotation shared across all tokens, we let adjacent frames share a rotation adapted to the content they have in common. Specifically, we divide the cache at each patch into temporal blocks of $n_B$ consecutive latent frames and fit one orthogonal rotation $R_B$ per block, which is applied to every frame in the block (Fig.~\ref{fig:blockwise-rotation}). Fitting the rotation separately for each block allows it to track the scene as it evolves, while sharing the rotation within a block amortizes its storage cost over $n_B$ cache vectors.

What should this shared rotation accomplish? An asymmetric quantizer stores a separate offset for each vector, allowing any component that is constant across channels to be represented without introducing additional quantization error. However, the temporal mean shared by nearby frames generally varies across channels, widening the range that must be quantized even when frame-specific deviations are small. We therefore seek a rotation $R_B$ that transforms the block's mean over time into a vector that is nearly constant across channels. This lets the quantizer's offset to capture the shared component, leaving its limited quantization levels to represent the smaller frame-specific deviations more accurately (we make this more concrete in Section \ref{sec:method}).

A general $d\times d$ rotation per block would be prohibitively expensive for a KV-cache quantized to two bits per scalar. To reduce this overhead, we construct each rotation as a balanced tree of $2\times2$ planar rotations followed by a fixed Hadamard transform shared across all blocks. Each planar rotation mixes two channels, with its angle chosen from $\{0,\pi/4,\pi/2,3\pi/4\}$. It is therefore stored with two bits and applied using only additions, subtractions, and a shared scaling. The tree uses $d-1$ such rotations to progressively concentrate the energy of the block's temporal mean into a single coordinate (Fig.~\ref{fig:rotation-tree}), which the Hadamard transform then spreads evenly across all $d$ channels. For $d=128$, the adaptive rotation parameters require only 254 bits per block.

We fit the angles greedily, level by level in the tree, without gradients or changes to model weights. For a fixed tree (which refers to the chosen pairings of channels for the planar rotations), this procedure is globally optimal for our mean-flattening objective (Theorem~\ref{thm:fixed-tree-optimality}). We also provide a custom Triton implementation of our scheme, which means that these rotations can be optimized online with almost no run-time overhead (Sec.~\ref{sec:kernels}). In experiments on LongCat-Video-13B, Self-Forcing-Wan-1.3B, and HY-WorldPlay-8B, \method{} outperforms the state-of-the-art video KV-cache quantization method QVG~\cite{qvg} at matched compression on all seven evaluated video fidelity and quality metrics for LongCat and Self-Forcing, and on six of seven for HY-WorldPlay (Sec.~\ref{sec:experiments}).
\vspace{-3mm}
\trqUseCompactHeadingSpacing
\section{Related Work}
\label{sec:related}
\paragraph{Diffusion and flow-based generation.}
Diffusion transformers replace convolutional denoisers with transformer backbones in latent generative models~\cite{dit}. LongCat-Video combines a diffusion-transformer architecture with video-continuation training and long-video inference~\cite{longcat}. HY-WorldPlay-8B is a streaming video diffusion world model that generates long-horizon video autoregressively in chunks, conditioned on user actions such as camera motion, and dynamically reconstitutes context from past frames to maintain long-term geometric consistency~\cite{worldplay2025}. Our method targets the conditioning KV-cache used during such long-horizon generation. It does not alter the training objective, sampling schedule, or model parameters.

\compactparagraph{Quantization of diffusion models.}
Q-Diffusion addresses post-training quantization of denoising networks through timestep-aware calibration and split shortcut quantization~\cite{qdiffusion}. Those mechanisms concern quantizing the denoiser activations and network weights. The goal of this work is to instead quantize the KV-cache which is reused over a continuation's denoising steps. The closest work is QVG~\cite{qvg}, which groups similar cache tokens, subtracts each group's centroid and quantizes the remaining residuals in several stages. QVG explicitly represents cache patterns using centroids. Our method instead represents a local temporal mean implicitly through an adaptive orthogonal transform, motivated by the success of orthogonal transforms for the quantization of LLMs \cite{spinquant, quarot, thinquant}.

\compactparagraph{Rotations and KV-cache quantization in LLMs.}
Early methods for KV-cache quantization in LLMs mainly exploit the distinct structure of keys and values. KIVI quantizes keys per channel and values per token~\cite{kivi}, and KVQuant finds that pre-RoPE keys are more friendly for quantization~\cite{kvquant}. A second line of work rotates the cache before quantization so that outlier channels are spread across many coordinates. QuaRot uses orthogonal rotations to redistribute activation outliers while folding the inverse transform into the weights~\cite{quarot}, SpinQuant learns rotations using Cayley SGD~\cite{spinquant}, and RotateKV applies them before RoPE and across grouped heads~\cite{rotatekv}. PolarQuant and TurboQuant use random rotations to make the coordinate distribution predictable enough for simple scalar or polar quantizers~\cite{polarquant,turboquant}. In all of these methods, the rotation is either fixed or calibrated once and shared across tokens, and it targets outliers within each vector. Our transform instead adapts to each temporal block and targets the mean shared by neighboring frames. Because these transforms vary across blocks, they cannot be absorbed into one global rotation, so we parameterize them in a much more light-weight fashion (Sec.~\ref{sec:planar}).

\section{Preliminaries}
\label{sec:prelim}
\subsection{Video continuation and conditioning caches}

Video diffusion transformers generate in the compressed latent space of a video
VAE, where each latent time index summarizes several RGB frames and each spatial
position corresponds to an image patch. Long videos are produced by
\emph{continuation}: rather than generating the whole video at once, the model
generates it chunk by chunk, each chunk a short sequence of latent frames
conditioned on everything generated so far. Let $\bm{x}_1,\ldots,\bm{x}_c$ denote the chunks
generated (or provided by the user) so far. The next chunk $\bm{x}^*=\bm{x}_{c+1}$ is
generated conditioned on the clean prefix
$\bm{x}^{\rm ctx}_c=(\bm{x}_1,\ldots,\bm{x}_c)$. Once $\bm{x}_{c+1}$ has been generated, it is appended to the prefix, and the
process repeats with $\bm{x}^{\rm ctx}_{c+1}=(\bm{x}_1,\ldots,\bm{x}_{c+1})$. We omit text
conditioning from the notation for simplicity.

LongCat-Video and the other models tested in this paper are trained with flow
matching~\cite{flowmatching}. With a time coordinate $t\in[0,1]$ running from data
($t=0$) to noise ($t=1$), training interpolates linearly between the target chunk
and Gaussian noise, $\bm{x}_t=(1-t)\bm{x}^*+t\,\bm{\epsilon}$ with $\bm{\epsilon}\sim\mathcal N(0,\bm{I})$.
The velocity along this path is $d\bm{x}_t/dt=\bm{\epsilon}-\bm{x}^*$.
During training, the ground-truth chunk $\bm{x}^*$ is known, and a transformer
$\bm{v}_\phi$ is trained to predict this velocity:
\begin{equation}
 \mathcal L(\phi)
 =
 \E_{(\bm{x}^{\rm ctx}_c,\bm{x}^*),\,\bm{\epsilon},\,t}
 \bigl[
 \|\bm{v}_\phi(\bm{x}_t,t,\bm{x}^{\rm ctx}_c)-(\bm{\epsilon}-\bm{x}^*)\|_2^2
 \bigr]
\end{equation}
Here the expectation is over context--target pairs, Gaussian noise, and the
training-time distribution over $t$.

At inference, generation proceeds in the reverse direction, from noise to data.
Each new chunk is initialized from noise, $\bm{x}^{(0)}_{c+1}\sim\mathcal N(0,\bm{I})$, and the
learned velocity field is approximately integrated from $t=1$ to $t=0$. For
instance, for timesteps $1=t_0>t_1>\cdots>t_J=0$,
inference applies denoising iteratively:
\begin{equation}
 \bm{x}^{(j+1)}_{c+1}
 =
 \bm{x}^{(j)}_{c+1}
 +(t_{j+1}-t_j)\,
 \bm{v}_\phi(\bm{x}^{(j)}_{c+1},t_j,\bm{x}^{\rm ctx}_c)
\end{equation}
Each step requires a full forward pass through all transformer layers, and the
cached context enters this computation at every layer. In LongCat-Video~\cite{longcatcode},
the context is processed independently of the noisy chunk and its timestep, so
these keys and values are identical at every denoising step. They are therefore
computed once and stored as a KV-cache
$\mathcal C_c=\{(\bm{K}^{(\ell)}_c,\bm{V}^{(\ell)}_c)\}_{\ell=1}^{L}$,
which is reused for all denoising steps of chunk $c+1$.

After video chunk $\bm{x}_{c+1}$ has been generated, its clean latents are passed through
the network once more to produce their keys and values, and these are appended
to the existing cache rather than recomputing it. The new rows come from the $T_{c+1}S$ tokens of the new chunk ($T_{c+1}$
latent frames of $S$ patches each). After $c$ chunks, the prefix contains
$N_c=S\sum_{i=1}^{c}T_i$ tokens. With $L$ layers, $n_{\rm head}$ heads, and $d$
channels per head, the cache then holds $2\,L\,n_{\rm head}\,N_c\,d$
scalars, so its size grows linearly with the number of chunks generated and,
for long videos, quickly dominates inference memory. Because the cache is append-only, each block is formed from frames that are
already cached, and is compressed once. 

\subsection{Uniform Quantization}
\label{sec:quantizer}
For a row vector $\bm{y}\in\R^d$, a $b$-bit uniform quantizer stores
an offset $m(\bm{y})$, a scale $\Delta(\bm{y})$, and one integer code $q_i(\bm{y})\in\{0,\ldots,2^b-1\}$ for each entry $i=1,\ldots,d$.
For KV-cache vectors quantized online, this is typically implemented
as asymmetric round-to-nearest (RTN) quantization:
\begin{equation}
\begin{aligned}
m(\bm{y}) &= \min_i y_i,\quad
\Delta(\bm{y}) = \tfrac{\max_i y_i-m(\bm{y})}{2^b-1},\\
\widehat y_i &= m(\bm{y})+\Delta(\bm{y})q_i(\bm{y}),\quad
q_i(\bm{y}) = \operatorname{round}\bigl(\tfrac{y_i-m(\bm{y})}{\Delta(\bm{y})}\bigr).
\end{aligned}
\label{eq:affine}
\end{equation}
A key property is translation invariance: for
$\bm{y}'=\bm{y}+c\mathbf{1}_d^\top$, we have
$m(\bm{y}')=m(\bm{y})+c$,
$\Delta(\bm{y}')=\Delta(\bm{y})$, and
$\bm{q}(\bm{y}')=\bm{q}(\bm{y})$.
Thus, $\hat{\bm{y}}'=\hat{\bm{y}}+c\mathbf{1}_d^\top$,
so the quantization error is unchanged by any value
that is constant across channels. Our method exploits this invariance, as we now describe.

% \subsection{Uniform Quantization}
% \label{sec:quantizer}
% For a row vector $\bm{y}\in\R^d$, a $b$-bit uniform quantizer stores
% the row minimum as an offset, a scale $\Delta$, and one integer code per entry. For KV-cache and other vectors which need to be quantized online, this is typically implemented as asymmetric round-to-nearest (RTN) quantization:
% \begin{equation}
% \begin{aligned}
%  \Delta&=\frac{\max_i y_i-\min_i y_i}{2^b-1},\:
%  q_i=\operatorname{round}\frac{y_i-\min_j y_j}{\Delta},\\
%  \widehat y_i&=\min_j y_j+\Delta q_i.
% \end{aligned}
% \label{eq:affine}
% \end{equation}
% The integers $q_i$ lie in $\{0,\ldots,2^b-1\}$, while the vector $\hat{\bm{y}}$ has the same range as $\bm{y}$. A key property of asymmetric quantization is translation invariance. That is, replacing $\bm{y}$
% with $\bm{y}+c\mathbf1_d^\top$ shifts the offset by $c$ and leaves $\Delta$ and the
% codes $\bm{q}$ unchanged. The quantization error therefore depends on the row's range, not
% on any component that is constant across channels. 

% Our method exploits this invariance. Within a block of adjacent frames, we attempt to make the per-channel mean approximately constant across channels, which leads to a small range while the exact mean does not influence the quantization error.

\section{Method}
\label{sec:method}
% \subsection{Flattening the shared temporal component}
Fix a layer $\ell$, an attention head $h$, and a point in generation after $c$
chunks. The cached keys and values of this head form matrices:
\begin{equation*}
 \bm{K}^{(\ell,h)},\,\bm{V}^{(\ell,h)}\in\R^{N\times d},
\end{equation*}
with one row per context token, where $N=N_c$ is the current number of cached
tokens. We describe TRQ for the values $\bm{V}^{(\ell,h)}$. The keys are handled
identically and independently, with their own rotations. For readability, we drop
the layer and head superscripts and the chunk index from here on. The $N=TS$ context tokens come from $T=\sum_{i=1}^{c}T_i$ frames, each
divided into $S$ image patches. 

Consider the tokens covering the same image patch
in $n_B$ consecutive frames. Stacking their value vectors as rows gives a
temporal block $\label{eq:values}
 \bm{V}_B\in\R^{n_B\times d}$. Nearby frames typically show much of
the same visual content in a given patch, so their cached vectors
share a substantial component that is stable over time (Fig.~\ref{fig:kv-cache}).
Collecting these stable components in each channel to a vector gives:
\begin{equation*}
 \bm{\mu}_B=(\mu_1,\mu_2,\ldots,\mu_d)
 =\frac1{n_B}\mathbf1_{n_B}^\top \bm{V}_B
\end{equation*}
Writing each entry of $\bm{V}_B$ as its channel
mean plus a frame-specific deviation $\epsilon_{t,j}$, the block is:
\begin{equation*}
\bm{V}_B=
\begin{pmatrix}
\mu_1+\epsilon_{1,1} & \mu_2+\epsilon_{1,2} & \cdots & \mu_d+\epsilon_{1,d}\\
\mu_1+\epsilon_{2,1} & \mu_2+\epsilon_{2,2} & \cdots & \mu_d+\epsilon_{2,d}\\
\vdots & \vdots & \ddots & \vdots\\
\mu_1+\epsilon_{n_B,1} & \mu_2+\epsilon_{n_B,2} & \cdots & \mu_d+\epsilon_{n_B,d}
\end{pmatrix},
\end{equation*}
or equivalently, in compact form:
\begin{equation}
 \bm{V}_B=\mathbf1_{n_B}\bm{\mu}_B+\bm{E}_B,
\label{eq:decomp}
\end{equation}
 How large and stable the shared
component $\bm{\mu}_B$ is of course depends on how fast the motion of the content is, but we expect that, especially for small block sizes, we should see the residual component being small. The shared component accounts for most of the cache's squared norm,
particularly for short blocks (see Section~7.3 of the appendix). The shared term $\mathbf1_{n_B}\bm{\mu}_B$ is identical in every row but generally varies across
channels.
Unlike the constant shifts absorbed
by the quantizer's offset in Sec.~\ref{sec:quantizer}, this variation
can widen each row's range, increasing the quantization scale
$\Delta$ even when the frame-specific deviations are small.
We therefore change the \emph{channel coordinates}. Let
$\bm{R}_B\in\R^{d\times d}$ be an orthogonal transform ($\bm{R}_B \bm{R}^{\top}_B = \bm{I}_d$) applied to every row. Then:
\begin{equation}
 \bm{V}_B\bm{R}_B=\mathbf1_{n_B}(\bm{\mu}_B\bm{R}_B)+\bm{E}_B\bm{R}_B
\end{equation}
Ideally, we could choose
$\bm{R}_B$ so that:
\begin{equation}
 \bm{\mu}_B\bm{R}_B\approx c_B\mathbf1_d^\top,
\end{equation}
for some constant $c_B \in \mathbb{R}$, in which case:
\begin{equation*}
\bm{V}_B\bm{R}_B \approx
c_B\mathbf{1}_{n_B}\mathbf{1}_d^\top
+\bm{E}_B\bm{R}_B
\end{equation*}
% \begin{equation*}
%  \bm{V}_B\bm{R}_B\approx
%  c_B\begin{pmatrix}
%  1 & 1 & \cdots & 1\\
%  1 & 1 & \cdots & 1\\
%  \vdots & \vdots & \ddots & \vdots\\
%  1 & 1 & \cdots & 1
%  \end{pmatrix}
%  +\bm{E}_B\bm{R}_B
% \end{equation*}
Instead of a different shared value $\mu_j$ in each channel, the shared component
now contributes the same value $c_B$ to every channel and every frame. The
quantizer's offset absorbs it, and the quantization levels are spent on the
frame-specific deviations $\bm{E}_B\bm{R}_B$. This is similar in motivation to QVG's
centroid-based approach~\cite{qvg}, which separates representative cache patterns
from their residuals. Rather than explicitly subtracting and storing a centroid,
however, we change coordinates so that the quantizer's existing offset captures the
shared temporal component.

\subsection{Concentrating the channel means}

To make the temporal mean
approximately constant across channel, we construct $\bm{R}_B$ in two steps. We factor
$\bm{R}_B=\bm{G}_B\bm{H}$, where an adaptive orthogonal transform
$\bm{G}_B$ first concentrates the mean's mass into its first
coordinate, and a fixed Walsh--Hadamard transform $\bm{H}$ then
spreads that coordinate evenly across all channels. Only
$\bm{G}_B$ is adapted to the block, while $\bm{H}$ is the same
for every block. The key intuition here is that $\bm{H}$ can flatten
a vector with mass
concentrated in its first coordinate, but not an arbitrary vector.
We explain this property and how it motivates our construction.
\vspace{-3mm}
\paragraph{Flattening.}
For $d$ a power of two, let $\bm{H}$ be the normalized
Walsh--Hadamard matrix~\cite{walsh1923}. Its first row is
$\mathbf1_d^\top/\sqrt{d}$, so any vector with all its mass
in the first coordinate is mapped to a constant vector:
$(\sqrt{d}\,c,0,\ldots,0)\bm{H}=c\mathbf1_d^\top$.
However, applying $\bm{H}$ directly to an arbitrary mean vector
need not make it flatter. For example, it maps $(3,4,5,6)$
to $(9,-1,-2,0)$, increasing rather than reducing its range.
Indeed, the remaining rows of $\bm{H}$ contain positive and
negative entries and are orthogonal to the constant vector,
so mass outside the first coordinate produces variation
across channels. Thus, to exploit the flattening property
of $\bm{H}$, we first need to concentrate the temporal mean
of the block into its first coordinate. Once the mean is concentrated,
the spreading operation is independent of the block, so the
same $\bm{H}$ can be reused across all blocks.
\vspace{-5mm}
\paragraph{Concentrating.}
Motivated by this property, we choose an adaptive orthogonal
transform $\bm{G}_B$ that moves as much of $\bm{\mu}_B$ as
possible into the first coordinate. Ideally,
$\bm{\mu}_B\bm{G}_B=(\sqrt{d}\,c_B,0,\ldots,0)$, so that
$\bm{\mu}_B\bm{R}_B=c_B\mathbf1_d^\top$ is exactly constant.
In general, some mass remains in the other coordinates,
which we collect in a residual vector
$\bm{\rho}\in\mathbb{R}^{d-1}$:
\begin{equation}
\bm{\mu}_B\bm{G}_B=(\sqrt{d}\,c_B,\bm{\rho}^\top)
\end{equation}
Applying $\bm{H}$ then gives:
\begin{equation}
\bm{\mu}_B\bm{R}_B
=c_B\mathbf1_d^\top+(0,\bm{\rho}^\top)\bm{H}
\end{equation}
The first term is constant across channels and can be absorbed
by the quantizer's offset, while the second captures the
remaining variation of the temporal mean. Because the rows
of $\bm{H}$ are orthogonal, this residual has zero channel
mean and preserves the norm $\|\bm{\rho}\|_2$. Therefore,
\begin{equation}
\|\bm{\mu}_B\bm{R}_B-c_B\mathbf1_d^\top\|_2^2
=\|\bm{\rho}\|_2^2
=\|\bm{\mu}_B\|_2^2-dc_B^2
\label{eq:identity}
\end{equation}
Thus, concentrating more mass into the first coordinate
directly reduces the variation of the transformed mean
across channels. An exact concentrating transform always exists, for example
a Householder reflection~\cite{householder1958}, but
it requires a full-precision $d$-dimensional vector per block.
We instead construct $\bm{G}_B$ from a tree of discrete planar
rotations requiring only $2(d-1)$ bits (Sec.~\ref{sec:planar}).

\subsection{Tree of discrete planar rotations}
\label{sec:planar}
Since we are concerned with two (or very low) bit quantization of the KV-cache,
every piece of additional metadata is important. A block of $n_B$ cache vectors is
stored with only $2n_Bd$ bits of codes, e.g., 2048 bits for $n_B=8$ and $d=128$. A
Householder vector stored in 16-bit precision would cost another 2048 bits,
doubling the storage of the block. The transform $\bm{G}_B$ also needs to be applied to
every cached row at encoding and inverted at decoding, so it must be cheap to
apply as well. We therefore need a concentrating transform that is described by
very few bits and can be applied with a handful of additions/subtractions per channel.
\vspace{-5mm}
\paragraph{Planar rotations.}
We build $\bm{G}_B$ from the smallest rotations available, planar (Givens)
rotations~\cite{givens1958}, which mix only two channel means at a time. Each such
planar rotation right-multiplies a pair of coordinates by:
\begin{equation}
 \bm{Q}(\theta)=\begin{pmatrix}\cos\theta&-\sin\theta\\\sin\theta&\cos\theta\end{pmatrix},
\end{equation}
so that a pair of mean coordinates $(\mu_1,\mu_2)$ becomes
$(z(\theta),r(\theta))=(\mu_1\cos\theta+\mu_2\sin\theta,-\mu_1\sin\theta+\mu_2\cos\theta)$.
Since $\bm{Q}(\theta)$ is orthogonal, $z(\theta)^2+r(\theta)^2=\mu_1^2+\mu_2^2$, so
maximizing $z(\theta)^2$ is equivalent to minimizing
$r(\theta)^2$.

% Because $\bm{Q}(\theta)$ is orthogonal, the two outputs have the
% same total squared norm $\mu_1^2+\mu_2^2$ as the inputs, so maximizing the squared norm of the
% first coordinate minimizes that left in the residual.
For example, $\theta=\pi/4$ computes a normalized sum and difference. Writing
$\mu_1=m-\delta$ and $\mu_2=m+\delta$ makes its effect clear:
\begin{equation*}
 (m-\delta,\,m+\delta)\longmapsto(\sqrt2\,m,\,\sqrt2\,\delta).
\end{equation*}
The first coordinate captures the common component and the residual captures the
difference, which vanishes when the inputs are equal. Thus $\pi/4$ suits pairs of
similar values with the same sign, but not every pair. To adapt to different
magnitudes and signs, we choose each angle from
$\Theta=\{0,\pi/4,\pi/2,3\pi/4\}$, which gives four mappings:
\begin{equation}
\begin{aligned}
 \theta&=0: &(\mu_1,\,\mu_2)&\longmapsto(\mu_1,\,\mu_2),\\
 \theta&=\pi/2: &(\mu_1,\,\mu_2)&\longmapsto(\mu_2,\,-\mu_1),\\
 \theta&=\pi/4: &(m-\delta,\,m+\delta)&\longmapsto(\sqrt2\,m,\,\sqrt2\,\delta),\\
 \theta&=3\pi/4: &(-m+\delta,\,m+\delta)&\longmapsto(\sqrt2\,m,\,-\sqrt2\,\delta)
\end{aligned}
\label{eq:four-angles}
\end{equation}
Each choice suits a different kind of pair. The choice $\theta=0$ forwards $\mu_1$
unchanged and suits pairs in which the first channel already dominates,
$|\mu_1|\gg|\mu_2|$, leaving only the small $\mu_2$ in the residual. The swap
$\theta=\pi/2$ sends $\mu_2$ to the first coordinate and suits the opposite case, $|\mu_2|\gg|\mu_1|$.
The normalized sum $\theta=\pi/4$ suits similar values with the same sign, where the
common component $m$ is large and the difference $\delta$ small. The normalized
difference $\theta=3\pi/4$ suits similar magnitudes with opposite signs, for which
the sum would cancel but the difference preserves the shared magnitude $m$. In each
case, the residual is small exactly when the pair matches the pattern the angle is
designed for. Since every direction in the plane lies within $22.5^\circ$ of one of
these four axes, the best of the four angles always concentrates at least $\cos^2(22.5^\circ)\approx85\%$ of the pair's mass on the first coordinate. Each angle is stored with two bits, and applying a
planar rotation requires no general multiplications, only additions/subtractions, or a rescaling.

For each pair of channels with current mean coordinates $(\mu_1,\mu_2)$, we select:
\begin{equation}
 \theta^\star=\argmax_{\theta\in\Theta}
 (\mu_1\cos\theta+\mu_2\sin\theta)^2
\label{eq:angle}
\end{equation}
The first coordinate always carries at least
as much mass as either input alone, $(\mu_1\cos\theta^\star+\mu_2\sin\theta^\star)^2
\geq\max\{\mu_1^2,\mu_2^2\}$.

\vspace{-5mm}

\begin{figure}[t]
\centering
\newcommand{\tgate}[5]{% #1 row in, #2 row out, #3 forwarded slot, #4 residual slot, #5 angle
  \node[rotation] (g#1-#3) at ($(c#1-#3)!0.5!(c#2-#4)$) {$#5$};
  \draw[inp] (c#1-#3.south) -- (g#1-#3);
  \draw[inp] (c#1-#4.south) -- (g#1-#3);
  \draw[fwd] (g#1-#3) -- (c#2-#3.north);
  \draw[res] (g#1-#3) -- (c#2-#4.north);
}
\resizebox{\columnwidth}{!}{%
\begin{tikzpicture}[
    x=1.5cm, y=1cm,
    cell/.style={draw, minimum width=1.2cm, minimum height=0.6cm,
                 inner sep=0pt, font=\small},
    act/.style={cell, draw=cvprblue, fill=cvprblue!12},
    root/.style={cell, draw=cvprblue, fill=cvprblue!35, very thick},
    frz/.style={cell, draw=black!20, fill=black!4, text=black!40},
    flat/.style={cell, draw=orange!75!black, fill=orange!15},
    rotation/.style={draw=cvprblue, fill=white, rounded corners=2pt,
                 font=\scriptsize, inner sep=1.5pt,
                 minimum width=0.75cm, minimum height=0.4cm},
    inp/.style={cvprblue!60, thick},
    fwd/.style={cvprblue, thick, -{Stealth[length=4pt]}},
    res/.style={black!40, dashed, -{Stealth[length=4pt]}},
    lab/.style={font=\small, anchor=east},
    lvl/.style={font=\footnotesize, text=black!60, anchor=east},
    note/.style={font=\footnotesize, text=black!60, anchor=west},
]
  % channel indices
  \foreach \i in {1,...,4}
    \node[font=\scriptsize, text=black!45] at (\i-1, 0.5) {\i};

  % coordinates after each level
  \foreach \v/\s [count=\i] in {5/act,-4/act,4/act,5/act}
    \node[\s] (c0-\i) at (\i-1, 0) {$\v$};
  \foreach \v/\s [count=\i] in {-\tfrac{9}{\sqrt2}/act,-\tfrac{1}{\sqrt2}/frz,
                                \tfrac{9}{\sqrt2}/act,\tfrac{1}{\sqrt2}/frz}
    \node[\s] (c1-\i) at (\i-1, -1.45) {$\v$};
  \foreach \v/\s [count=\i] in {9/root,-\tfrac{1}{\sqrt2}/frz,0/frz,\tfrac{1}{\sqrt2}/frz}
    \node[\s] (c2-\i) at (\i-1, -2.9) {$\v$};
  \foreach \v [count=\i] in {4.50,4.50,3.79,5.21}
    \node[flat] (c3-\i) at (\i-1, -4.5) {$\v$};

  % planar rotations: level 1, 2
  \tgate{0}{1}{1}{2}{3\pi/4}
  \tgate{0}{1}{3}{4}{\pi/4}
  \tgate{1}{2}{1}{3}{3\pi/4}

  % Hadamard spreading
  \node[draw=orange!75!black, fill=orange!6, rounded corners=3pt,
        minimum width=5.7cm, minimum height=0.5cm, font=\small]
        (H) at (1.5, -3.7) {$\bm{H}$: spread the root over all channels};
  \draw[fwd] (c2-1.south) -- (c2-1.south |- H.north);
  \foreach \i in {2,...,4}
    \draw[black!25, -{Stealth[length=3pt]}] (c2-\i.south) -- (c2-\i.south |- H.north);
  \foreach \i in {1,...,4}
    \draw[orange!75!black, thick, -{Stealth[length=4pt]}] (c3-\i.north |- H.south) -- (c3-\i.north);

  % row labels
  \node[lab] at (-0.6,  0)     {$\bm{\mu}_B$};
  \node[lvl] at (-0.6, -0.725) {level 1};
  \node[lvl] at (-0.6, -2.175) {level 2};
  \node[lab] at (-0.6, -2.9)   {$\bm{\mu}_B\bm{G}_B$};
  \node[lab] at (-0.6, -4.5)   {$\bm{\mu}_B\bm{G}_B\bm{H}$};
  \node[note] at (3.55,  0)    {range 9};
  \node[note] at (3.55, -2.9)  {$=(\sqrt d\,c_B,\bm{\rho}^{\top})$};
  \node[note] at (3.55, -4.5)  {range 1.41};
\end{tikzpicture}%
}
\caption{Tree of planar rotations for the four-channel mean
$\bm{\mu}_B=(5,-4,4,5)$ of Sec.~\ref{sec:planar}.}
\label{fig:rotation-tree}
\end{figure}
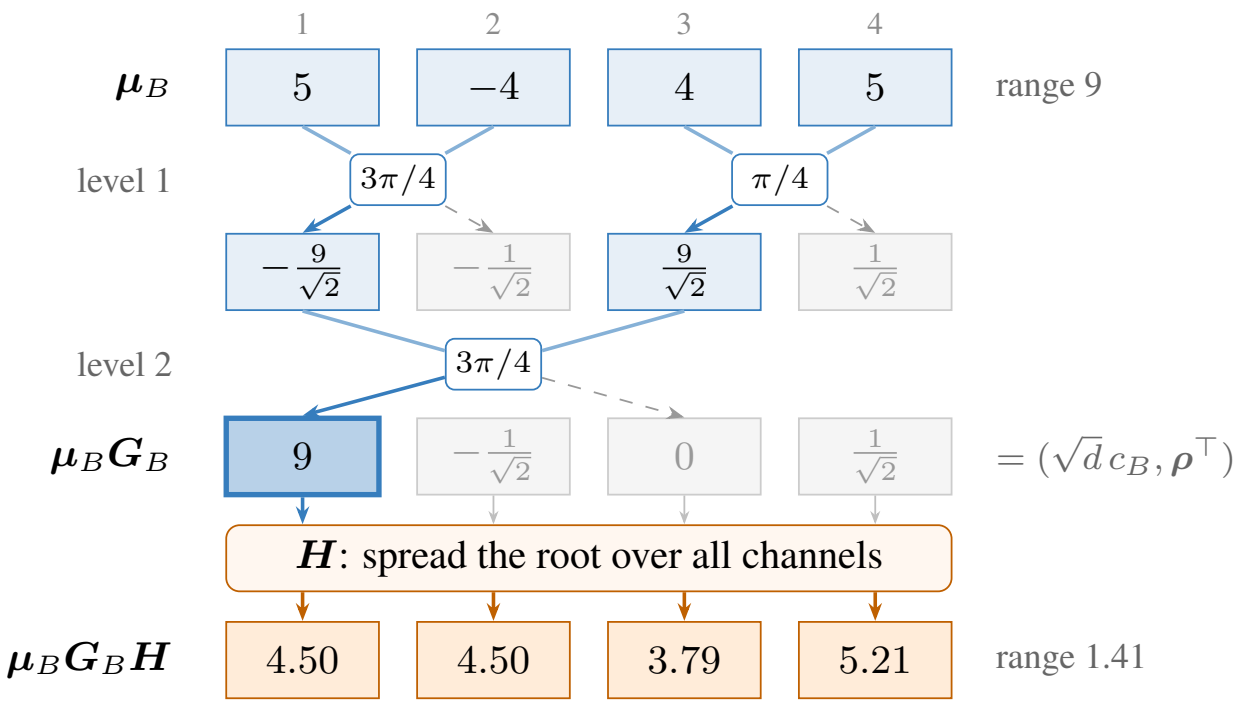

\paragraph{The tree.}
A single planar rotation concentrates only two coordinates, whereas the Hadamard
step requires the mass of all $d$ channel means to be concentrated in a single
coordinate. We therefore construct a binary tree to compose planar rotations and
combine all $d$ channel means into one root coordinate. Each planar rotation merges
two active coordinates into one forwarded coordinate, so exactly $d-1$ rotations are
needed to reduce $d$ active coordinates to a single root. We arrange them in a
balanced binary tree (see Fig.~\ref{fig:rotation-tree}). For $d=2^S$, the first level pairs
the $d$ channels; each subsequent level pairs the coordinates forwarded by the
previous one, halving the number of active coordinates. After $S=\log_2d$ levels,
a single root coordinate remains active, while all $d$ coordinates are retained in
the transform, the residual of every rotation stays in its own coordinate, so
nothing is discarded.

% The balanced arrangement has two advantages over, for example, a chain that
% merges one channel at a time. First, the rotations within a level act on disjoint
% pairs and can be applied in parallel. Second, every channel mean passes through only
% $\log_2d$ rotations on its way to the root rather than up to $d-1$, which limits how
% much squared norm can be lost along the way (Eq.~\eqref{eq:bound}).

Formally, a planar rotation acting on coordinates $(i,j)$ is the $d\times d$ matrix
$\bm{R}_{ij}(\theta)$ obtained by embedding $\bm{Q}(\theta)$ into $\bm{I}_d$ at rows and columns $i$
and $j$. Writing $\theta_{s,k}$ for the angle of the $k$-th rotation at level $s$ of the tree:
\begin{equation}
 \bm{L}_s=\prod_{k=0}^{d/2^s-1}\bm{R}_{i_{s,k}j_{s,k}}(\theta_{s,k}),\qquad
 \bm{G}_B=\bm{L}_1\bm{L}_2\cdots \bm{L}_S
 \label{eq:tree-formalization}
\end{equation}
Because vectors multiply from the left, the first level appears leftmost. Every
factor is orthogonal, so $\bm{G}_B^\top \bm{G}_B=\bm{I}_d$, and it is never necessary to
materialize a dense matrix. In total, $\bm{G}_B$ costs $2(d-1)$ bits, 254 bits for
$d=128$, or about 0.25 bits per scalar for $n_B=8$, and $O(n_Bd)$ operations to
apply.

The angles are fitted greedily from the leaves to the root. We initialize a working
vector with $\bm{\mu}_B$ and traverse the tree level by level where at each planar rotation,
we select the angle by Eq.~\eqref{eq:angle} and immediately update the two
coordinates. Fitting thus requires only the block mean and no gradients. Our Triton implementation fuses each tree level for fast \textit{online} fitting;
see Section~7.3 of the appendix.

 \compactparagraph{Example.}
Figure~\ref{fig:rotation-tree} illustrates the procedure for $\bm{\mu}_B=(5,-4,4,5)$. The tree progressively concentrates $81$ of the vector's total squared norm of $82$ into the first coordinate. The  Hadamard transform then spreads this mass across channels, reducing the range from $9$ to approximately $1.41$.

% \paragraph{Guarantee.}
% The forwarded direction is chosen from four axes, equivalently eight signed
% directions spaced by $45^\circ$. Therefore each planar rotation retains at least
% $\alpha=\cos^2(\pi/8)\approx0.8536$ of its input pair's mass in the forwarded
% coordinate. On a balanced tree, induction gives:
% \begin{equation}
%  d\,c_B^2\geq\alpha^{\log_2d}\|\bm{\mu}_B\|_2^2
% \label{eq:bound}
% \end{equation}

\compactparagraph{Applying the transform.}
Although the angles are fitted to the mean, the same $\bm{G}_B$ is applied to every frame
in the block:
\begin{equation*}
 \bm{V}_B\bm{G}_B=\mathbf1_{n_B}(\bm{\mu}_B\bm{G}_B)+\bm{E}_B\bm{G}_B
\end{equation*}
The transform acts on channels, not on time, so it remains $d\times d$ for any block
length, including a shorter final block. Only the shared mean is concentrated and the
frame-specific deviations are rotated along with it and quantized as usual.

Applying $\bm{R}_B=\bm{G}_B\bm{H}$ to a cache vector is inexpensive and never requires materializing a dense
$d\times d$ matrix. Each of the $d-1$ planar rotations, by
Eq.~\eqref{eq:four-angles}, either copies its inputs ($\theta=0$), swaps them with a
sign change ($\theta=\pi/2$), or forms a sum and a difference followed by a scaling
by $1/\sqrt2$ ($\theta\in\{\pi/4,3\pi/4\}$). The tree therefore costs at most
$2(d-1)$ additions per vector. The Hadamard step is computed with the fast
Walsh--Hadamard transform in $d\log_2d$ additions and subtractions. Its
normalization by $1/\sqrt d$ scales every channel equally and can be folded into the
quantizer's scale and offset. For $d=128$ (as in LongCat-Video-13B), the full transform thus takes roughly
$1{,}150$ additions/subtractions per vector, or about nine per channel. The rotations within each tree level and the butterflies within each Hadamard
stage act on disjoint pairs and can run in parallel, and every row of the block is
processed independently.

Decoding inverts the transform at the same cost. Since $\bm{H}$ is symmetric and
orthogonal, $\bm{R}_B^\top=\bm{H}\bm{G}_B^\top$. We apply the fast Walsh--Hadamard transform again
and then the planar rotations in reverse order with transposed angles, read from
the stored two-bit codes. This cost is small relative to the attention computation
in which the decoded vector is used: each cached key or value is read by every query
of the current chunk, which costs $O(d)$ operations per query, whereas the
transform is applied once per cached vector.

\subsection{Global optimality for a fixed tree}
\label{sec:optimality}
We now show that the greedy rule in Eq.~\eqref{eq:angle}
achieves the best possible mean flattening for a fixed
tree. A tree layout $T$ specifies the ordered channels pairs
$(i_{s,k},j_{s,k})$, indexed by level $s$ and rotation
position $k$ within that level, as defined in
Eq.~\eqref{eq:tree-formalization}. In each pair, the
first coordinate is forwarded to the next level,
while the second remains as a residual. The final
forwarded coordinate occupies position 1. The
pairings and order of the levels are fixed and only
the rotation angles vary.
Thus, with $\Theta=\{0,\pi/4,\pi/2,3\pi/4\}$, the feasible
set is:
\begin{equation}
\begin{aligned}
\mathcal G_T=\bigg\{
&\prod_{s=1}^{S}\prod_{k=0}^{d/2^s-1}
\bm R_{i_{s,k}j_{s,k}}(\theta_{s,k}):
\theta_{s,k}\in\Theta\ \text{for all }s,k
\bigg\}
\end{aligned}
\label{eq:feasible-tree}
\end{equation}
There are $4^{d-1}$ possible angle assignments to
this fixed tree.

For the block mean $\bm{\mu}_B$ and the normalized
Hadamard matrix $\bm H$, we define the mean-flatness
objective as:
\begin{equation}
\mathcal J_B(\bm G):=
\min_{c\in\mathbb R}
\|\bm{\mu}_B\bm G\bm H-c\mathbf1_d^\top\|_2^2
\label{eq:mean-flatness-objective}
\end{equation}
This measures the squared deviation of the transformed
mean from its best constant approximation. As seen in Eq. \eqref{eq:identity}, this is equivalent to minimizing the residual $\|\bm \rho \|_2^2$.

\begin{rotationtheorem}[Global optimality for a fixed tree]
\label{thm:fixed-tree-optimality}
For any block mean $\bm{\mu}_B$ and fixed tree layout
$T$, let
$\bm G_\star$ be obtained by selecting each angle
greedily according to Eq.~\eqref{eq:angle}, from leaves
to root. Then:
\begin{equation}
\mathcal J_B(\bm G_\star)
=\min_{\bm G\in\mathcal G_T}\mathcal J_B(\bm G)
\label{eq:global-mean-optimum}
\end{equation}
\end{rotationtheorem}

Given $\bm\mu_B$, greedy fitting therefore solves the joint
optimization over all $4^{d-1}$ angle assignments in $O(d)$
operations. Although the objective couples angles across levels,
for this angle set the best attainable parent magnitude depends
only on its input magnitudes and is nondecreasing in each.
Maximizing a child's forwarded magnitude therefore cannot worsen
the best outcome at its parent. We provide the proof in Section~7.1 of the appendix. To connect mean flatness to reconstruction error, the following corollary provides a bound on the quantization reconstruction error of our procedure.

\begin{rotationcorollary}[Quantization error bound]
\label{cor:affine-reconstruction}
Let $n_B,b\geq1$ be integers and
$\bm X_B=\mathbf1_{n_B}\bm\mu_B+\bm E_B\in\mathbb R^{n_B\times d}$,
where $\mathbf1_{n_B}^\top\bm E_B=\bm0$.
For each $\bm G\in\mathcal G_T$, set $\bm R=\bm G\bm H$ and
$\widehat{\bm X}_B(\bm G)=\mathcal Q_b(\bm X_B\bm R)\bm R^\top$.
Then:
\begin{equation}
\frac{\|\widehat{\bm X}_B(\bm G)-\bm X_B\|_F^2}{n_Bd}
\leq
\frac{\mathcal J_B(\bm G)+\|\bm E_B\|_F^2/n_B}
{2(2^b-1)^2}
\label{eq:affine-reconstruction-bound}
\end{equation}
Moreover, Theorem~\ref{thm:fixed-tree-optimality} implies that
greedy fitting globally minimizes the reconstruction-error
upper bound in Eq.~\eqref{eq:affine-reconstruction-bound}
over $\mathcal G_T$.
\end{rotationcorollary}

The bound in Eq.~\eqref{eq:affine-reconstruction-bound} identifies the
two contributions to reconstruction error, the frame-dependent variation
$\|\bm{E}_B\|_F^2/n_B$ and the residual from
$\mathcal J_B(\bm{G})$. If
$\mathcal J_B(\bm{G})=0$, translation invariance of
Eq.~\eqref{eq:affine} gives:
\begin{equation}
 \begin{aligned}
 \mathcal Q_b(\bm{X}_B\bm{R})-\bm{X}_B\bm{R}
 ={}&\mathcal Q_b(\bm{E}_B\bm{R})-\bm{E}_B\bm{R}
 \end{aligned}
 \label{eq:exact-mean-cancellation}
\end{equation}
Thus an exactly flattened mean contributes no
quantization error under asymmetric uniform quantization and only the rotated frame-specific deviations remain.
For the proof, see Section~7.2 of the appendix. %With approximate flattening, the guarantee of Theorem \ref{thm:fixed-tree-optimality} gives the smallest residual mean energy attainable by the chosen tree layout.

\subsection{Calibration, encoding, and metadata}
\compactparagraph{Offline tree layout selection.}
Since the channels in $\bm{K}$ or $\bm{V}$ can be reordered arbitrarily without changing the full-precision output, we use a small calibration process to determine which channels meet in the tree (i.e. the tree layout $T$). We represent this layout by a permutation $\bm{P}$, selected
independently for each layer, head, and key, value. To determine this, we
use eight videos disjoint from evaluation. The permutation search starts with the original layout and proposes 32 swaps of
subtrees, accepting changes that reduce the squared reconstruction error
on the training blocks. The calibration videos determine
the shared layout, but increasing their number does not change the online
algorithm or layout size. Once selected, $\bm{P}$ is fixed across evaluation videos. In contrast, the angles are
fitted to each new block by Eq.~\eqref{eq:angle}, which optimizes mean flatness for
new videos.
Unlike the fitted angles, $\bm{H}$ is determined entirely by $d$ and
requires neither calibration nor per-block metadata.

\begin{table*}[t]
\centering
\caption{Generated-video quality and encoded cache compression on 100 matched videos per method at 480p and 25 matched videos per method at 720p. The model rows give BF16 VBench scores. Bold values indicate the best quantized results within each model and resolution at the reported precision, including ties. Uncertainties denote sample standard errors across videos.}
\lbltbl{results}
% \vspace{-2.5mm}
\setlength{\tabcolsep}{4pt}
\renewcommand{\arraystretch}{0.85}
\resizebox{0.95\textwidth}{!}{%
\begin{tabular}{l|c|c|ccc|cccc}
\toprule
\makecell[bl]{\textbf{Method}} & \makecell[b]{\textbf{Compression}\\\textbf{Ratio (BF16)}} & \makecell[b]{\textbf{Runtime}\\\textbf{/ Dense}$\downarrow$} & \makecell[b]{\textbf{PSNR}$\uparrow$} & \makecell[b]{\textbf{SSIM}$\uparrow$} & \makecell[b]{\textbf{LPIPS}$\downarrow$} & \makecell[b]{\textbf{Background}\\\textbf{Consistency}$\uparrow$} & \makecell[b]{\textbf{Image}\\\textbf{Quality}$\uparrow$} & \makecell[b]{\textbf{Subject}\\\textbf{Consistency}$\uparrow$} & \makecell[b]{\textbf{Aesthetic}\\\textbf{Quality}$\uparrow$} \\
\midrule
\multicolumn{2}{l}{\textit{LongCat-Video-13B}} & 1.00$\times$ & \multicolumn{3}{l}{\textit{INT2 KV-cache, 480p}} & $94.61\pm0.30$ & $67.95\pm0.89$ & $93.02\pm0.57$ & $63.68\pm0.99$ \\
\midrule
TurboQuant & $\mathbf{7.64}\times$ & $1.05\times$ & $19.40\pm0.34$ & $0.66\pm0.01$ & $0.31\pm0.01$ & $88.50\pm0.46$ & $43.44\pm1.26$ & $81.36\pm0.88$ & $48.53\pm1.01$ \\
QuaRot & $7.53\times$ & $1.03\times$ & $22.71\pm0.46$ & $0.79\pm0.01$ & $0.16\pm0.01$ & $93.08\pm0.35$ & $62.85\pm1.00$ & $89.93\pm0.64$ & $59.25\pm0.96$ \\
KIVI & $7.37\times$ & $\mathbf{1.02}\times$ & $23.18\pm0.44$ & $0.81\pm0.01$ & $0.15\pm0.01$ & $93.22\pm0.37$ & $62.03\pm1.10$ & $90.16\pm0.72$ & $60.32\pm1.03$ \\
QVG & $6.88\times$ & $1.07\times$ & $24.06\pm0.46$ & $0.83\pm0.01$ & $0.13\pm0.01$ & $93.30\pm0.31$ & $63.46\pm0.95$ & $90.38\pm0.58$ & $60.50\pm0.94$ \\
\rowcolor{cyan!20} \textbf{\method{}} & $6.89\times$ & $1.05\times$ & $\mathbf{26.53}\pm0.57$ & $\mathbf{0.87}\pm0.01$ & $\mathbf{0.10}\pm0.01$ & $\mathbf{94.10}\pm0.32$ & $\mathbf{65.46}\pm0.90$ & $\mathbf{92.02}\pm0.57$ & $\mathbf{61.98}\pm0.95$ \\
\midrule
\multicolumn{2}{l}{\textit{LongCat-Video-13B}} & 1.00$\times$ & \multicolumn{3}{l}{\textit{INT2 KV-cache, 720p}} & $94.86\pm0.54$ & $67.18\pm1.85$ & $93.96\pm0.96$ & $61.33\pm2.34$ \\
\midrule
TurboQuant & $\mathbf{7.64}\times$ & $1.03\times$ & $18.86\pm0.67$ & $0.65\pm0.03$ & $0.36\pm0.02$ & $89.98\pm0.95$ & $43.94\pm2.41$ & $82.75\pm2.06$ & $46.99\pm2.29$ \\
QuaRot & $7.53\times$ & $1.00\times$ & $22.15\pm0.87$ & $0.77\pm0.03$ & $0.19\pm0.02$ & $92.99\pm0.79$ & $63.84\pm2.01$ & $90.68\pm1.36$ & $58.40\pm2.43$ \\
KIVI & $7.47\times$ & $\mathbf{0.97}\times$ & $22.44\pm0.83$ & $0.79\pm0.03$ & $0.19\pm0.02$ & $93.80\pm0.63$ & $61.92\pm2.18$ & $91.70\pm1.21$ & $59.19\pm2.42$ \\
QVG & $7.12\times$ & $1.06\times$ & $24.50\pm0.96$ & $0.83\pm0.02$ & $0.15\pm0.02$ & $93.35\pm0.74$ & $64.49\pm1.92$ & $91.06\pm1.12$ & $59.19\pm2.31$ \\
\rowcolor{cyan!20} \textbf{\method{}} & $6.90\times$ & $1.05\times$ & $\mathbf{25.59}\pm1.04$ & $\mathbf{0.85}\pm0.02$ & $\mathbf{0.13}\pm0.02$ & $\mathbf{94.25}\pm0.71$ & $\mathbf{65.97}\pm1.79$ & $\mathbf{93.46}\pm1.06$ & $\mathbf{61.11}\pm2.44$ \\
\midrule
\multicolumn{2}{l}{\textit{HY-WorldPlay-8B}} & 1.00$\times$ & \multicolumn{3}{l}{\textit{INT2 KV-cache, 480p}} & $91.65\pm0.44$ & $64.71\pm1.06$ & $88.08\pm0.76$ & $58.10\pm1.18$ \\
\midrule
TurboQuant & $\mathbf{3.28}\times$ & $1.09\times$ & $19.07\pm0.25$ & $0.61\pm0.01$ & $0.34\pm0.01$ & $87.69\pm0.35$ & $53.32\pm1.18$ & $80.16\pm0.60$ & $48.15\pm0.89$ \\
QuaRot & $3.27\times$ & $1.06\times$ & $22.03\pm0.28$ & $0.74\pm0.01$ & $0.22\pm0.01$ & $89.76\pm0.41$ & $62.37\pm1.02$ & $85.05\pm0.69$ & $54.81\pm1.01$ \\
KIVI & $3.25\times$ & $\mathbf{1.03}\times$ & $23.90\pm0.32$ & $0.82\pm0.01$ & $0.16\pm0.01$ & $91.20\pm0.40$ & $62.93\pm1.06$ & $\mathbf{87.61}\pm0.66$ & $56.27\pm1.11$ \\
QVG & $3.15\times$ & $1.06\times$ & $21.16\pm0.29$ & $0.71\pm0.01$ & $0.24\pm0.01$ & $90.49\pm0.42$ & $\mathbf{64.16}\pm1.04$ & $86.33\pm0.68$ & $56.00\pm1.04$ \\
\rowcolor{cyan!20} \textbf{\method{}} & $3.19\times$ & $1.08\times$ & $\mathbf{27.50}\pm0.41$ & $\mathbf{0.90}\pm0.01$ & $\mathbf{0.09}\pm0.01$ & $\mathbf{91.40}\pm0.43$ & $63.44\pm1.09$ & $\mathbf{87.61}\pm0.74$ & $\mathbf{57.38}\pm1.16$ \\
\midrule
\multicolumn{2}{l}{\textit{Self-Forcing-Wan-1.3B}} & 1.00$\times$ & \multicolumn{3}{l}{\textit{INT2 KV-cache, 480p}} & $92.26\pm0.50$ & $65.59\pm1.03$ & $87.94\pm1.06$ & $60.29\pm1.11$ \\
\midrule
TurboQuant & $\mathbf{5.30}\times$ & $1.08\times$ & $17.99\pm0.42$ & $0.58\pm0.02$ & $0.38\pm0.01$ & $89.01\pm0.51$ & $56.17\pm1.23$ & $81.70\pm1.03$ & $52.91\pm1.02$ \\
QuaRot & $5.25\times$ & $1.05\times$ & $20.81\pm0.42$ & $0.70\pm0.01$ & $0.22\pm0.01$ & $91.09\pm0.48$ & $62.83\pm1.12$ & $85.68\pm0.98$ & $56.94\pm1.00$ \\
KIVI & $5.24\times$ & $\mathbf{1.02}\times$ & $18.76\pm0.39$ & $0.59\pm0.02$ & $0.36\pm0.01$ & $87.48\pm0.42$ & $55.13\pm1.26$ & $78.85\pm0.84$ & $50.47\pm0.89$ \\
QVG & $4.98\times$ & $1.12\times$ & $22.45\pm0.51$ & $0.75\pm0.01$ & $0.18\pm0.01$ & $91.85\pm0.49$ & $64.81\pm1.09$ & $87.14\pm1.06$ & $59.63\pm1.10$ \\
\rowcolor{cyan!20} \textbf{\method{}} & $5.01\times$ & $1.10\times$ & $\mathbf{25.37}\pm0.65$ & $\mathbf{0.83}\pm0.01$ & $\mathbf{0.12}\pm0.01$ & $\mathbf{92.31}\pm0.48$ & $\mathbf{65.37}\pm1.01$ & $\mathbf{88.11}\pm1.02$ & $\mathbf{60.23}\pm1.09$ \\
\bottomrule
\end{tabular}%
}
% \vspace{2mm}
\end{table*}

\subsection{Integration with attention}
\label{sec:attention}
We apply the rotation when the model constructs or adds to its conditioning
cache. For each temporal block, the rotated keys and values are
$\bm{Y}_{K,B}=\bm{K}_B\bm{R}_{K,B}$ and $\bm{Y}_{V,B}=\bm{V}_B\bm{R}_{V,B}$.
The key and value rotations use independently fitted angles,
but the same $\bm{H}$ is shared by both and across all blocks.
After quantization and dequantization, let
$\widehat{\bm{Y}}_{K,B}$ and $\widehat{\bm{Y}}_{V,B}$ denote the corresponding reconstructed
tensors in the rotated channel coordinates. Before attention, we map them back
to the model's original channel coordinates,
$\widehat{\bm{K}}_B=\widehat{\bm{Y}}_{K,B}\bm{R}_{K,B}^{\top}$ and
$\widehat{\bm{V}}_B=\widehat{\bm{Y}}_{V,B}\bm{R}_{V,B}^{\top}$.
Since $\bm{R}_B$ is orthogonal, its inverse exactly recovers
the original cache vectors in the absence of quantization.
Thus, the rotation changes only the coordinates presented
to the quantizer, leaving the network's full-precision
computation unchanged. After undoing the calibrated channel
permutation, the reconstructed keys and values enter the
model's existing attention operation. 

\subsection{Efficient Triton Kernel}\label{sec:kernels}
We implement our block-wise rotation scheme with specialized Triton kernels for CUDA GPUs. We fuse the rotation fitting operation for all tree levels into one kernel, and each forward
or inverse traversal loads a tile of channel vectors once and writes the result
after the final level. Independent temporal blocks and spatial sites run in
parallel, avoiding the repeated indexing operations and intermediate tensors
of a level-by-level implementation.  These changes reduce kernel-launch and intermediate-memory overhead
without changing the quantization algorithm or stored representation. For timings against native PyTorch, see Section~7.3 of the appendix.

\section{Experiments}
\label{sec:experiments}

\subsection{Experimental setup}
Following QVG~\cite{qvg}, we evaluate LongCat-Video-13B, HY-WorldPlay-8B, and Self-Forcing-Wan-1.3B at
480p under the long-generation setting, comparing each quantized method against
the corresponding BF16 KV-cache baseline. We clone the QVG package and reproduce the models and their baseline directly from this. We follow \cite{qvg} and use the prompt suite from the
MovieGen Video Bench~\cite{moviegen}, specifically the extended prompt list
adopted by Self-Forcing. MovieGen Video Bench
contains 1,003 prompts spanning human activity, animals, natural scenes, physical
phenomena, and unusual subjects and activities. We sample 100 prompts from this collection, and use the same prompts across models and our baselines.
% \paragraph{Quantization and calibration.}
% \method{} uses eight-latent-frame temporal groups, INT2 codes,
% and scaled FP4 E2M1 minima and scales in 64-channel groups.
% Permutation calibration and online angle fitting follow
% Sec.~\ref{sec:method}. Offline calibration uses eight videos
% with images and prompts disjoint from evaluation.
% For a fair comparison, QuaRot~\cite{quarot} and KIVI~\cite{kivi} use the same scale format as TRQ:
% INT2 codes with scaled FP4 E2M1 minima and scales in 64-channel groups.
% For QuaRot, we use the official implementation but quantize only the
% KV-cache. QVG~\cite{qvg} does not store asymmetric minima and scales in this form,
% so we keep their method as is, using the default settings from the official
% repository with INT2 cache quantization. We adapt TurboQuant~\cite{turboquant} to 64-channel groups with scaled
% FP4 E2M1 norm metadata, without calibration. We use one-bit MSE
% quantization plus one-bit QJL residual correction for keys and
% two-bit MSE quantization for values.
\compactparagraph{Quantization and calibration.}
\method{} uses eight-latent-frame temporal groups (i.e. $n_B=8$), INT2 codes,
and FP4 E2M1 minima and scales in 64-channel groups. We use these FP4 minima and scales to avoid extra storage overhead when the cache itself is quantized to 2-bits.
Permutation calibration and online angle fitting follow
Sec.~\ref{sec:method}. Offline calibration uses eight videos
with images and prompts disjoint from evaluation. We ablate the components of \method{} in Section~7.3 of the appendix.
For a fair comparison, QuaRot~\cite{quarot}, KIVI~\cite{kivi},
and TurboQuant~\cite{turboquant} use the same
INT2 code precision, group size of 64, and scaled FP4 E2M1
metadata precision as TRQ. QuaRot and KIVI store minima and scales,
while TurboQuant stores norms.
For QuaRot, we use the official implementation but quantize only the
KV-cache. QVG~\cite{qvg} does not store asymmetric minima and scales in this form,
so we keep their method as is, using the default settings from the official
repository with INT2 cache quantization.
TurboQuant requires no calibration; we use one-bit MSE quantization
plus one-bit QJL residual correction for keys and two-bit MSE
quantization for values.
% \vspace{-3mm}
\compactparagraph{Metrics.}
We follow the evaluation protocol of QVG \cite{qvg} and evaluate PSNR, SSIM, and LPIPS on the first 20 raw decoded RGB frames with the paired
BF16 output. For LongCat these are the first continuation chunks; for HY and Self-Forcing these
are zero-based frames 29--48 and 93--112 which occur after the first compressed context is used, respectively.
PSNR is averaged per frame in decibels, SSIM uses the repository's $11\times11$
box window with zero padding, and LPIPS uses RGB inputs in $[0,1]$ with
normalization enabled. VBench background consistency, image quality, subject consistency, and aesthetic quality are reported as percentages using CLIP, MUSIQ, DINO, and LAION respectively as evaluators, as in \cite{qvg}.
Their inputs are the encoded 180-frame continuations for LongCat, the
complete 285-frame HY videos, and 700 frames for
Self-Forcing. Thus the three fidelity metrics measure the first affected
window, whereas VBench evaluates the entire rollout window.

\subsection{Generated-video quality}
Table~\ref{tab:results} compares \method{} with TurboQuant~\cite{turboquant}, QuaRot~\cite{quarot}, KIVI~\cite{kivi} and QVG~\cite{qvg} at 2-bit
precision. On LongCat-Video and Self-Forcing-Wan, \method{} achieves the best
score on all seven fidelity and VBench metrics, at both 480p and 720p for
LongCat. On HY-WorldPlay, it is best or tied on six of seven metrics; the
exception is image quality, where QVG scores higher. These improvements carry over to the full
rollouts evaluated by VBench. Subject consistency, which reflects long-range
coherence, improves by 1.6 points over QVG on LongCat and 1.0 point on
Self-Forcing and \method{} significantly narrows the gap to the BF16 dense model.

\section{Conclusion and Limitations}
\label{sec:conclusion}

We introduced TRQ a 2-bit KV-cache
quantization method for video diffusion transformers. TRQ exploits the
observation that adjacent frames in a video cache share a slowly varying
per-channel mean. It fits a separate rotation to each temporal block that
flattens this mean, so the quantizer's offset can absorb the shared component.
Each rotation is a tree of $2\times2$ planar rotations with two-bit angles,
followed by a fixed Hadamard transform. Across
three video diffusion transformers, TRQ substantially improves fidelity over QVG and other KV-cache compression baselines at similar
compression levels.
TRQ benefits from exploiting temporal redundancy. This leaves open how best to
handle content where that redundancy is weaker, such as fast motion or scene
cuts. Block lengths that adapt to scene dynamics are a natural way to address
this. Moreover, TRQ and centroid-based methods such as QVG exploit
complementary structure, and combining the two could lead to gains over either alone. Finally, the online rotation fitting explored in TRQ has not been done to our knowledge for other model families and modalities, and is a promising avenue for future work.

% \trqRestoreHeadingSpacing
% \pagebreak 
{\fontsize{9}{10.5}\selectfont
\bibliographystyle{ieeenat_fullname}
\bibliography{main}
}

\clearpage
\trqRestoreHeadingSpacing
\section{Appendix}

\subsection{Proof of Theorem \ref{thm:fixed-tree-optimality}}
\label{sec:proof_tree_thm}

\begin{proof}
% Fix the row-vector mean $\bm{\mu}_B\in\mathbb{R}^{1\times d}$,
% the binary tree layout $T$, and the normalized Hadamard matrix
% $\bm{H}$. The admissible angle set is
% $\Theta=\{0,\pi/4,\pi/2,3\pi/4\}$, and $\mathcal{G}_T$ is the
% set of transforms generated by all assignments of these angles
% to the internal nodes of $T$.

% We use the tree convention in the construction: each internal
% node receives the two coordinates forwarded by its children,
% applies one planar rotation, and forwards the first output to
% its parent. The second output remains as a residual coordinate
% and is not used by any subsequent tree rotation. The final
% root node is stored in the first coordinate before applying
% $\bm{H}$, as in the decomposition
% $\bm{\mu}_B\bm{G}=(z,\bm{\rho}^{\top})$.
% All residual coordinates are retained, so every
% $\bm{G}\in\mathcal{G}_T$ is a product of embedded orthogonal
% planar rotations and is itself orthogonal.

% We first show that minimizing the mean-flatness objective is
% equivalent to maximizing the squared root output. We then prove
% that the greedy procedure attains the largest possible root
% output magnitude over all angle assignments on $T$.

Fix the block mean $\bm{\mu}_B$ and tree layout $T$.
Every $\bm G\in\mathcal G_T$ is orthogonal, since it
is a product of rotations. We first show that
minimizing our objective Eq~\eqref{eq:mean-flatness-objective} is equivalent to maximizing
the squared root output of the tree, and then prove that the greedy algorithm attains the largest possible output magnitude
over all angle assignments on $T$. Consider an arbitrary $\bm{G}\in\mathcal{G}_T$ and write:
\[
 \bm{\mu}_B\bm{G}=(z,\bm{\rho}^{\top}),
 \qquad \bm{\rho}\in\mathbb{R}^{d-1}
\]
Let $\bm{e}_1$ be the first standard basis column vector, then:
\[
 \bm{e}_1^\top\bm{H}
 =\frac{\mathbf{1}_d^\top}{\sqrt d}
\]
Transposing this equality and multiplying by $\bm{H}$, using
$\bm{H}\bm{H}^\top=\bm{I}_d$, gives:
\[
 \bm{H}\mathbf{1}_d=\sqrt d\,\bm{e}_1
\]
Thus,
\[
 \bm{\mu}_B\bm{G}\bm{H}
 =\frac{z}{\sqrt d}\mathbf{1}_d^\top+\bm{w},
 \qquad
 \bm{w}:=(0,\bm{\rho}^\top)\bm{H}
\]
The residual row $\bm{w}$ has zero channel sum, since:
\[
 \begin{aligned}
 \bm{w}\mathbf{1}_d
 &=(0,\bm{\rho}^\top)\bm{H}\mathbf{1}_d\\
 &=\sqrt d\,(0,\bm{\rho}^\top)\bm{e}_1=0
 \end{aligned}
\]

This comes from the property of the normalized Walsh-Hadamard matrix that all rows except the first have entries summing to zero \cite{ailon2006approximate}. 
Moreover, orthogonality of $\bm{H}$ implies
$\|\bm{w}\|_2^2=\|\bm{\rho}\|_2^2$ since an orthogonal transform does not change the norm.
Thus $\bm{w}$ is orthogonal to every constant row vector.
For any $c\in\mathbb{R}$, expanding the norm gives:
\begin{equation}
 \begin{aligned}
 &\|\bm{\mu}_B\bm{G}\bm{H}
       -c\mathbf{1}_d^\top\|_2^2\\
 &\quad=\left\|\bm{w}
       +\left(\frac{z}{\sqrt d}-c\right)
        \mathbf{1}_d^\top\right\|_2^2\\
 &\quad=\|\bm{w}\|_2^2
       +2\left(\frac{z}{\sqrt d}-c\right)
        \bm{w}\mathbf{1}_d
       +d\left(\frac{z}{\sqrt d}-c\right)^2\\
 &\quad=\|\bm{\rho}\|_2^2
       +d\left(c-\frac{z}{\sqrt d}\right)^2
 \end{aligned}
 \label{eq:proof-tree-offset-decomposition}
\end{equation}
The second term is nonnegative and becomes zero at
$c=z/\sqrt d$. Therefore this is the unique minimizer, and:
\[
 \mathcal{J}_B(\bm{G})=\|\bm{\rho}\|_2^2
\]
Because $\bm{G}$ is orthogonal,
\[
 \|\bm{\mu}_B\|_2^2
 =\|\bm{\mu}_B\bm{G}\|_2^2
 =z^2+\|\bm{\rho}\|_2^2
\]
It follows that, for every transform $\bm{G}$,
\begin{equation}
 \mathcal{J}_B(\bm{G})
 =\|\bm{\mu}_B\|_2^2-z^2
 \label{eq:proof-tree-root-objective}
\end{equation}
The first term is fixed. Hence minimizing
$\mathcal{J}_B$ is exactly equivalent to maximizing $z^2$,
or equivalently $|z|$, over all angle assignments on $T$. This completes the first part of the proof.

\noindent We now prove the main theorem statement, that the greedy algorithm minimizes the objective across all transforms $\bm{G} \in \mathcal{G}_T$. Consider a node of the binary rotation tree, corresponding to a planar rotation that combines two coordinates. Let $a,b\in\mathbb R$ denote its two inputs, which are either original channel means at the first level or coordinates forwarded from the previous level of the tree. The coordinate forwarded by a rotation with angle $\theta$ is:
\[
 f_\theta(a,b):=a\cos\theta+b\sin\theta
\]
For the four angles, its possible values are:
\[
 \begin{array}{c|c}
 \theta & f_\theta(a,b)\\ \hline
 0 & a\\
 \pi/4 & (a+b)/\sqrt2\\
 \pi/2 & b\\
 3\pi/4 & (b-a)/\sqrt2
 \end{array}
\]
Set $A=|a|$ and $B=|b|$. The sum and difference options
satisfy:
\[
 \begin{aligned}
 \max\{(a+b)^2,(b-a)^2\}
 &=a^2+b^2+2|ab|\\
 &=(|a|+|b|)^2
 \end{aligned}
\]
Taking square roots therefore gives:
\[
 \max\{|a+b|,|b-a|\}=A+B
\]
Thus the largest magnitude is:
\begin{equation}
 \begin{aligned}
 \max_{\theta\in\Theta}|f_\theta(a,b)|
 &=F(A,B),\\
 F(A,B)&:=\max\left\{
     A,\ B,\ \frac{A+B}{\sqrt2}
   \right\}
 \end{aligned}
 \label{eq:monotone-rotation}
\end{equation}

This holds for every sign combination of $a$ and
$b$. In particular, if $ab\geq0$, the angle $\pi/4$ attains $(A+B)/\sqrt2$; if $ab<0$, the angle $3\pi/4$
attains that magnitude. The angles $0$ and $\pi/2$ attain
$A$ and $B$, respectively. Hence it
depends only on the input magnitudes, even though the angle
that achieves it may depend on the input signs.

The function $F$ is nondecreasing in each argument on
$[0,\infty)^2$. Indeed, if $0\leq A\leq A'$ and
$0\leq B\leq B'$. Therefore
\begin{equation}
 F(A,B)\leq F(A',B')
 \label{eq:proof-tree-monotonicity}
\end{equation}
These two properties, dependence only on magnitudes and
monotonicity, allow locally optimal choices to be composed
throughout the tree. For each node $v$, let $T_v$ denote the subtree rooted at $v$.
Define $\mathcal{S}_v \subset \mathbb{R}$ as the set of all values
that node $v$ can forward to its parent, considering every possible
assignment of rotation angles within $T_v$. Define:
\[
 A_v:=\max_{s\in\mathcal{S}_v}|s|
\]
Each set $\mathcal{S}_v$ is finite and nonempty, so this
maximum is attained. We prove the claim by induction on the structure of the
binary rotation tree, starting at the leaves (i.e. before any rotation has been performed and the channel means $\bm{\mu}_B$ take their original values) and proceeding
toward the root.

\noindent At an original leaf node $i$ there are no prior angles to choose, 
thus:
\[
 \mathcal{S}_i=\{\mu_{B,i}\},
 \qquad A_i=|\mu_{B,i}|
\]

Now let $v$ be an internal node with children $\ell$ and $r$.
The two child subtrees have disjoint leaves and disjoint
angle variables. Since the residual coordinates do not participate in
subsequent rotations, each child passes only its
forwarded scalar to node $v$.  Any admissible assignment in $T_\ell$
can therefore be combined with any assignment in
$T_r$, giving
\[
 \mathcal{S}_v
 =\left\{
 f_\theta(a,b):
 \begin{array}{l}
 a\in\mathcal{S}_\ell,\ b\in\mathcal{S}_r,\\
 \theta\in\Theta
 \end{array}
 \right\}
\]
We claim that the exact subtree optimum is:
\begin{equation}
 A_v=F(A_\ell,A_r)
 \label{eq:proof-tree-magnitude-recurrence}
\end{equation}
We prove both the upper bound and attainability.

\emph{Upper bound.}
Take arbitrary child outputs
$a\in\mathcal{S}_\ell$ and $b\in\mathcal{S}_r$, and an
arbitrary parent angle $\theta\in\Theta$.

By definition, $|a|\leq A_\ell$ and $|b|\leq A_r$.
Using Eq.~\eqref{eq:monotone-rotation} and the monotonicity of $F$,
we obtain
\[
\begin{aligned}
|f_\theta(a,b)|
&\leq \max_{\varphi\in\Theta}|f_\varphi(a,b)|\\
&=F(|a|,|b|)\\
&\leq F(A_\ell,A_r)
\end{aligned}
\]

This holds for every admissible assignment in $T_v$,
so $A_v\leq F(A_\ell,A_r)$.

\emph{Attainability.}
Choose assignments in the two child subtrees whose outputs
$a_\star$ and $b_\star$ satisfy
$|a_\star|=A_\ell$ and $|b_\star|=A_r$.
Such assignments exist and, by subtree independence, can be
used simultaneously. Choose an admissible angle at $v$
that maximizes $|f_\theta(a_\star,b_\star)|$.
Equation~\eqref{eq:monotone-rotation} then gives:
\[
 \begin{aligned}
 |f_{\theta_\star}(a_\star,b_\star)|
 &=F(|a_\star|,|b_\star|)\\
 &=F(A_\ell,A_r)
 \end{aligned}
\]
This is an attainable output of $T_v$, regardless of the
signs of $a_\star$ and $b_\star$. Thus
$A_v\geq F(A_\ell,A_r)$, proving the claimed equality.

We now connect this recurrence to the greedy rule in
Eq.~\eqref{eq:angle}. At a leaf $i$, the actual forwarded
value is $\mu_{B,i}$ and already has magnitude $A_i$.
Suppose inductively that greedy fitting in the subtrees of
an internal node $v$ produces outputs $a_g$ and $b_g$ with:
\[
 |a_g|=A_\ell,\qquad |b_g|=A_r
\]
At $v$, the greedy rule maximizes $f_\theta(a_g,b_g)^2$.
Maximizing this square is equivalent to maximizing the
absolute value, so the magnitude forwarded by the selected
rotation is
\[
 \begin{aligned}
 \max_{\theta\in\Theta}|f_\theta(a_g,b_g)|
 &=F(|a_g|,|b_g|)\\
 &=F(A_\ell,A_r)=A_v
 \end{aligned}
\]
The induction therefore holds at $v$. Proceeding from the
leaves to the root proves that greedy fitting attains $A_v$
at every node and, in particular, the largest possible
root magnitude $A_{\rm root}$.

Now, for any $\bm G\in\mathcal G_T$, let $z$ denote its
root output. By definition, $|z|\leq A_{\rm root}$,
while greedy fitting attains $|z_\star|=A_{\rm root}$.
Therefore, by Eq.~\eqref{eq:proof-tree-root-objective},
\[
\begin{aligned}
\mathcal J_B(\bm G)
&=\|\bm{\mu}_B\|_2^2-z^2\\
&\geq\|\bm{\mu}_B\|_2^2-A_{\rm root}^2\\
&=\mathcal J_B(\bm G_\star)
\end{aligned}
\]
Since $\bm G_\star\in\mathcal G_T$, it attains the
global minimum, proving the theorem.

\end{proof}

\subsection{Proof of Corollary \ref{cor:affine-reconstruction}}
\label{sec:proof-affine-reconstruction}

\begin{proof}
Fix any $\bm G\in\mathcal G_T$ and use
$\bm R=\bm G\bm H$ as in the corollary. Since both factors
are orthogonal, $\bm R$ is orthogonal as well.

For the row quantizer in Eq.~\eqref{eq:affine}, rounding
to the nearest integer incurs an error of at most half
clof the quantization step in each coordinate. Thus, for any row $\bm y$,
\begin{equation}
 \|\mathcal Q_b(\bm y)-\bm y\|_2^2
 \leq \frac{d(\max_i y_i-\min_i y_i)^2}{4(2^b-1)^2}
 \label{eq:proof-affine-rounding-error}
\end{equation}

For any $c\in\mathbb R$, subtracting $c$ from every
coordinate leaves the range unchanged. For a
row that is not constant, the maximum and minimum occur at distinct coordinates. The squared difference of their deviations from $c$ is at
most twice the sum of their squared deviations. Hence:
\[
 (\max_i y_i-\min_i y_i)^2
 \leq 2\sum_{i=1}^d(y_i-c)^2
\]
 Combining it with
Eq.~\eqref{eq:proof-affine-rounding-error} gives:
\begin{equation}
 \|\mathcal Q_b(\bm y)-\bm y\|_2^2
 \leq \frac{d}{2(2^b-1)^2}
       \|\bm y-c\mathbf1_d^\top\|_2^2
 \label{eq:range-to-energy}
\end{equation}

\noindent We apply this inequality to every row of $\bm X_B\bm R$,
using the same scalar $c$ for all rows. Orthogonality of
$\bm R$ does not change the norm of ther reconstruction error, thus:
\[
 \begin{aligned}
 &\|\widehat{\bm X}_B(\bm G)-\bm X_B\|_F^2\\
 &\quad=\|\mathcal Q_b(\bm X_B\bm R)-\bm X_B\bm R\|_F^2\\
 &\quad\leq\frac{d}{2(2^b-1)^2}
   \|\bm X_B\bm R-c\mathbf1_{n_B}\mathbf1_d^\top\|_F^2
 \end{aligned}
\]
Using Eq.~\eqref{eq:decomp} for $\bm X_B$ gives:
\begin{equation}
 \begin{aligned}
 &\bm X_B\bm R-c\mathbf1_{n_B}\mathbf1_d^\top\\
 &\quad=\mathbf1_{n_B}(\bm\mu_B\bm R-c\mathbf1_d^\top)
       +\bm E_B\bm R
 \end{aligned}
 \label{eq:proof-affine-transformed-decomposition}
\end{equation}
Taking the squared Frobenius norm in
Eq.~\eqref{eq:proof-affine-transformed-decomposition}, the cross term is zero
because $\mathbf1_{n_B}^\top\bm E_B=\bm0$.
Also, $\|\bm E_B\bm R\|_F^2=\|\bm E_B\|_F^2$ by
orthogonality. Therefore
\begin{equation}
 \begin{aligned}
 &\|\bm X_B\bm R-c\mathbf1_{n_B}\mathbf1_d^\top\|_F^2\\
 &\quad=n_B\|\bm\mu_B\bm R-c\mathbf1_d^\top\|_2^2
       +\|\bm E_B\|_F^2
 \end{aligned}
 \label{eq:block-centered-energy}
\end{equation}
Since the bound holds for every $c\in\mathbb R$, minimize
over $c$ and use the definition of $\mathcal J_B$ in
Eq.~\eqref{eq:mean-flatness-objective}. Dividing by $n_Bd$
gives
\begin{equation}
 \begin{aligned}
 &\frac{\|\widehat{\bm X}_B(\bm G)-\bm X_B\|_F^2}{n_Bd}\\
 &\quad\leq
 \frac{n_B\mathcal J_B(\bm G)+\|\bm E_B\|_F^2}
      {2n_B(2^b-1)^2}\\
 &\quad=
 \frac{\mathcal J_B(\bm G)+\|\bm E_B\|_F^2/n_B}
      {2(2^b-1)^2}
 \end{aligned}
 \label{eq:proof-affine-arbitrary-transform-bound}
\end{equation}
Finally, $b$, $n_B$, and $\bm E_B$ are fixed, so this
upper bound depends on $\bm G$ only through
$\mathcal J_B(\bm G)$. Theorem~\ref{thm:fixed-tree-optimality}
therefore implies that greedy fitting globally minimizes
the bound.
\end{proof}

% Keep the tall channel figure in the two-column flow so it can occupy
% the spare column beside the end of the proofs.
% Do not stretch the short final proof column to the full page height.
\raggedbottom
\begin{figure}[H]
    \centering
    \includegraphics[width=\linewidth,height=0.88\textheight,keepaspectratio]{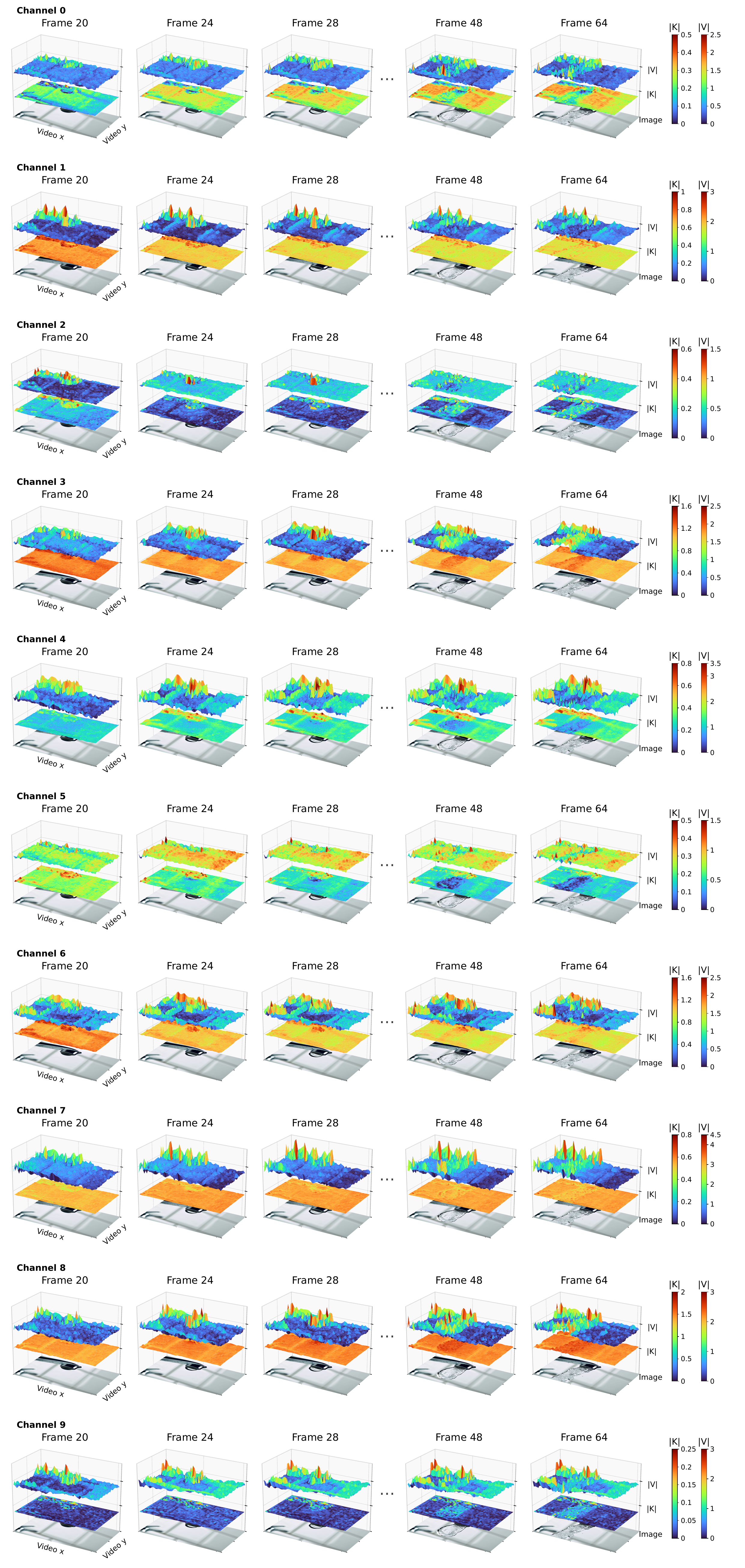}
    \caption{Figure \ref{fig:kv-cache} shown for an additional 10 channels. There exist differences \textit{across channels}, but within a channel adjacent frames tend to have similar values.}
    \label{fig:additional_channels}
\end{figure}

% The remaining ablation tables need the full page width.
\onecolumn
\flushbottom
\thispagestyle{plain}
\subsection{Additional Results and Ablations}

\begin{table}[H]
\centering
\caption{Ablation results on 20 matched videos per configuration at 480p.
VBench scores are percentages. Bold indicates the best result within each model.}
\label{tab:ablations}
\small
\setlength{\tabcolsep}{2pt}
\renewcommand{\arraystretch}{0.9}
\begin{tabular}{l|ccc|cccc}
\toprule
Config.
& PSNR$\uparrow$ & SSIM$\uparrow$ & LPIPS$\downarrow$
& BG$\uparrow$ & Img.$\uparrow$
& Subj.$\uparrow$ & Aesth.$\uparrow$ \\
\midrule
\multicolumn{8}{l}{\textit{LongCat-Video-13B}} \\
\midrule
\quad Full method
& \textbf{27.69} & \textbf{0.883} & \textbf{0.082}
& \textbf{94.60} & \textbf{69.04} & \textbf{94.21} & \textbf{63.06} \\
\quad $\bm{H}$ only
& 25.90 & 0.853 & 0.099
& 94.37 & 68.52 & 93.84 & 62.59 \\
\quad $\bm{H}$ + random $2{\times}2$
& 23.30 & 0.803 & 0.140
& 93.69 & 64.07 & 92.12 & 60.89 \\
\midrule
\multicolumn{8}{l}{\textit{HY-WorldPlay-8B}} \\
\midrule
\quad Full method
& \textbf{27.11} & \textbf{0.898} & \textbf{0.078}
& \textbf{91.56} & \textbf{65.80} & \textbf{88.53} & \textbf{59.36} \\
\quad $\bm{H}$ only
& 23.14 & 0.792 & 0.158
& 90.80 & 64.88 & 87.89 & 57.97 \\
\quad $\bm{H}$ + random $2{\times}2$
& 22.25 & 0.761 & 0.179
& 90.59 & 64.92 & 87.37 & 57.48 \\
\bottomrule
\end{tabular}
\end{table}

\begin{table}[H]
    \centering
    \caption{
        Mean and residual squared norm in LongCat KV caches,
        averaged over 10 videos across all 32 heads and 48 layers.
        This indicates the percentage of total cache mass
        within contiguous temporal blocks.
    }
    \label{tab:temporal_residual}
    \small
    \renewcommand{\arraystretch}{1.15}
    \begin{tabular}{@{}lrrrr@{}}
        \toprule
        & \multicolumn{2}{c}{\textbf{Keys}}
        & \multicolumn{2}{c}{\textbf{Values}} \\
        \cmidrule(lr){2-3} \cmidrule(l){4-5}
        \textbf{Block size ($n_B$)}
        & Mean ($\bm{\mu}_B$) & Residual ($\bm{E}_B$)
        & Mean ($\bm{\mu}_B$) & Residual ($\bm{E}_B$) \\
        \midrule
         2 & 91.22 &  8.78 & 77.59 & 22.41 \\
         4 & 85.23 & 14.77 & 64.92 & 35.08 \\
         8 & 80.61 & 19.39 & 57.08 & 42.92 \\
        16 & 76.88 & 23.12 & 51.94 & 48.06 \\
        \bottomrule
    \end{tabular}
\end{table}

\label{sec:runtime}
\begin{table}[H]
\centering
\small
\setlength{\tabcolsep}{4pt}
\caption{Complete timings for the online rotations on an NVIDIA H100 80GB. The custom Triton kernel substantially reduces the time required for multiplying and decoding the rotations.}
\lbltbl{timing}
\begin{tabular}{lrr}
\toprule
Implementation & \makecell{K/V pair\\(ms)} & \makecell{Total Cache\\(s)} \\
\midrule
Native PyTorch implementation & 6.638 & 10.196 \\
Eager Cuda graph & 3.174 & 4.876 \\
Compact CUDA graph (Triton) & 1.785 & 2.743 \\
\bottomrule
\end{tabular}
\label{tab:cuda_timings}
\end{table}

\begin{figure}[H]
    \centering
    \includegraphics[width=0.5\linewidth]{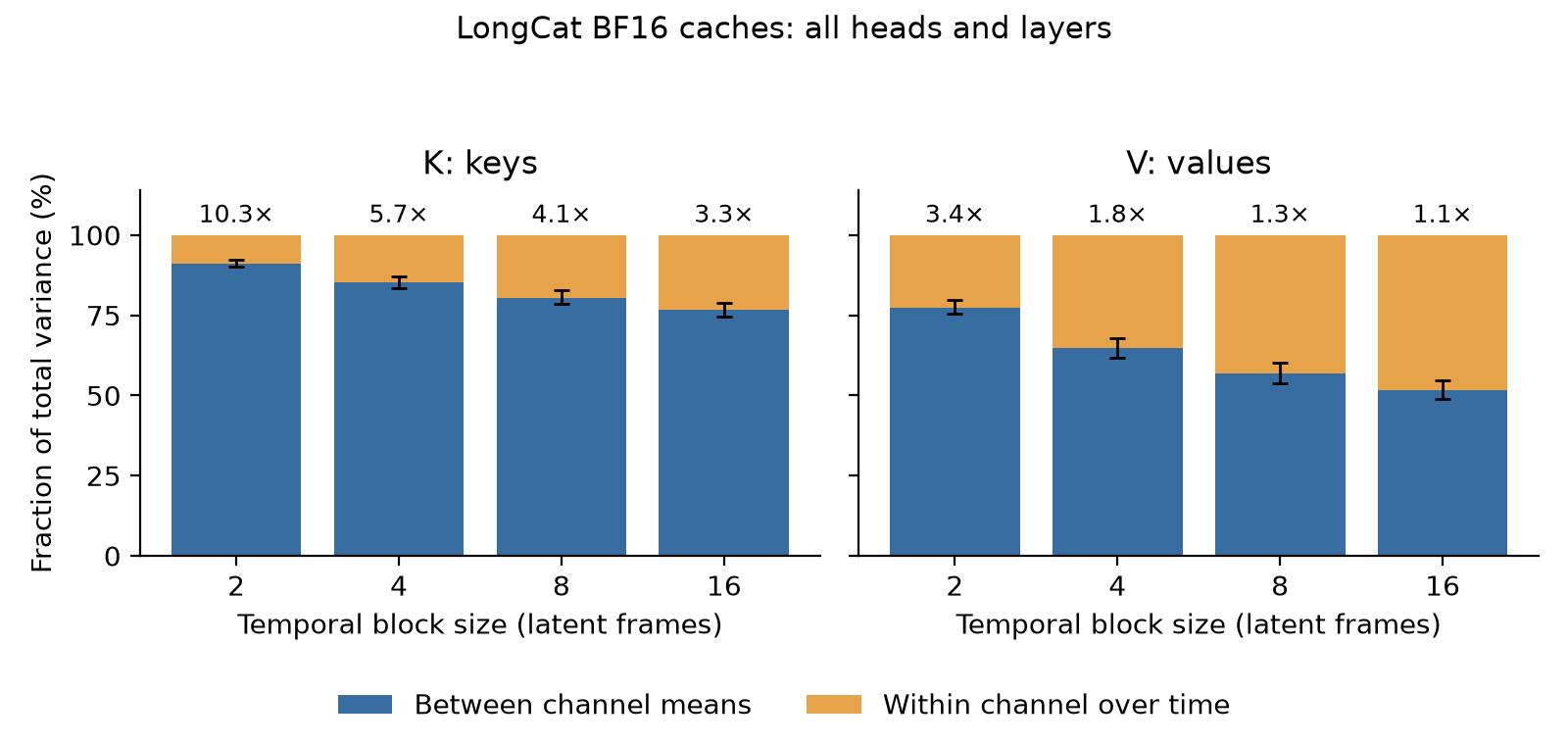}
    \caption{Variance across channels versus within a channel over time.}
    \label{fig:placeholder}
\end{figure}

\end{document}